\documentclass[authoryear,final,5p,times,twocolumn]{elsarticle}

\usepackage{appendix}

\usepackage{amssymb}
\usepackage{amsmath}
\usepackage{amsthm}
\usepackage{amsfonts}
\usepackage{mathrsfs}
\usepackage{newtxmath}
\usepackage{bm}
 
\usepackage{float} 
\usepackage{subfigure}
\usepackage{graphicx} 
\graphicspath{{./Figs/}{./figures/pdf/}{./figures/pdf/piecewise/}{./figures/pdf/piecewise/QPP_2/}{./figures/pdf/piecewise/QPP_1/}{./figures/pdf/example/}{./figures/pdf/valid/}{./figures/pdf/acc/}} 

\usepackage{enumerate}

\usepackage{makecell}
\usepackage{multirow} 
\usepackage{booktabs}
\newcommand{\tabincell}[2]{\begin{tabular}{@{}#1@{}}#2\end{tabular}}
\usepackage{footnote}
\makesavenoteenv{tabular}
\usepackage{threeparttable}

\usepackage{lscape}
\usepackage{color, soul}
\usepackage{ulem}
\usepackage{verbatim} 

\usepackage[misc]{ifsym}
\usepackage{overpic}

\usepackage[T1]{fontenc} 

\usepackage[ruled]{algorithm2e} 
\usepackage{color,xcolor}

\SetCommentSty{mycommfont}
\newcommand{\mathcolorbox}[2]{\colorbox{#1}{$\displaystyle #2$}}
\newcommand{\hltwo}[1]{{\sethlcolor{myhlcolortwo}\hl{#1}}}
\newcommand{\hlthree}[1]{{\sethlcolor{myhlcolorthree}\hl{#1}}}
\newcommand{\hlfour}[1]{{\sethlcolor{myhlcolorfour}\hl{#1}}}

\usepackage{algorithmic}

\usepackage[nameinlink]{cleveref}
\crefname{algocf}{alg.}{algs.}
\Crefname{line}{Algorithm}{Algorithms}

\gdef\ie{\textit{i.e.}}
\gdef\eg{\textit{e.g.}}
\gdef\etc{\textit{etc}}

\definecolor{matlabblue}{rgb}{0,0.4470,0.7410}
\definecolor{matlabred}{rgb}{0.8500,0.3250,0.0980}
\definecolor{matlabyellow}{rgb}{0.9290,0.6940,0.1250}
\definecolor{matlabpurple}{rgb}{0.4940,0.1840,0.5560}
\definecolor{matlabgreen}{rgb}{0.4660,0.6740,0.1880}
\definecolor{matlabcyan}{rgb}{0.3010,0.7450,0.9330}
\definecolor{matlabmagenta}{rgb}{0.6350,0.0780,0.1840}

\definecolor{myhlcolor}{rgb}{1, 1, 1}
\definecolor{myhlcolortwo}{rgb}{1, 1, 1}
\definecolor{myhlcolorthree}{rgb}{1, 1, 1}
\definecolor{myhlcolorfour}{rgb}{1, 1, 1}

\gdef\problem{difficulties in \textit{effective solution of multi-classification} and \textit{fast model selection}}

\newtheorem{theorem}{Theorem}

\newtheorem{definition}{Definition}
\newtheorem{corollary}{Corollary}

\def\mytitle{Solution Path Algorithm for Twin Multi-class Support Vector Machine} 
\def\myjournal{Expert Systems With Applications} 

\journal{\myjournal}

\soulregister{\allowbreak}{0}
\makeatletter
\AtBeginDocument{\def\@citecolor{cyan}}
\AtBeginDocument{\def\@urlcolor{magenta}}
\AtBeginDocument{\def\@linkcolor{blue}}
\makeatother

\begin{document}
\setulcolor{blue} 
\setstcolor{red} 
\sethlcolor{myhlcolor} 

\begin{frontmatter}

\title{\mytitle}


\author[addr0]{Liuyuan Chen\corref{cor0}}
\author[addr1]{Kanglei Zhou\corref{cor0}}
\author[addr2]{Junchang Jing}
\author[addr3]{Haiju Fan}
\author[addr4]{Juntao Li\corref{cor1}}
\ead{juntaolimail@126.com}

\cortext[cor0]{Equal contribution}
\cortext[cor1]{Corresponding author}

\address[addr0]{Journal Editorial Department, Henan Normal University, Xinxiang 453007, China.}

\address[addr1]{School of Computer Science and Engineering, Beihang University, Beijing 100191, China.}

\address[addr2]{Henan International Joint Laboratory of Cyberspace Security Applications, Information Engineering College, Henan University of Science and Technology, Luoyang 471023, China.}

\address[addr3]{College of Computer and Information Engineering, Henan Normal University, Xinxiang 453007, China.}

\address[addr4]{College of Mathematics and Information Science, Henan Normal University, Xinxiang 453007, China.}

\begin{abstract}
The twin support vector machine and its extensions have made great achievements in dealing with binary classification problems. However, it suffers from \problem. This work devotes to the fast regularization parameter tuning algorithm for the twin multi-class support vector machine. Specifically, a novel sample data set partition strategy is first adopted, which is the basis for the model construction. Then, combining the linear equations and block matrix theory, the Lagrangian multipliers are proved to be piecewise linear \textit{w.r.t.} the regularization parameters, so that the regularization parameters are continuously updated by only solving the break points. Next, Lagrangian multipliers are proved to be 1 as the regularization parameter approaches infinity, thus, a simple yet effective initialization algorithm is devised. Finally, eight kinds of events are defined to seek for the starting event for the next iteration. Extensive experimental results on nine UCI data sets show that the proposed method can achieve comparable classification performance without solving any quadratic programming problem.
\end{abstract}


\end{frontmatter}


\section{Introduction} \label{Introduction}
As a machine learning method for pattern classification, the well-known support vector machine (SVM) by solving a quadratic programming problem (QPP) has shown the great prospect and excellent generalization performance after decades of evolutionary development since it was proposed by \cite{VapnikSupport}. 
Based on the structural risk minimization principle and Vapnik-Chervonenkis dimensional theory in statistical learning theory, SVM has been widely used in data mining \citep{LuoDataMining, VaidyaPrivacy}, knowledge discovery \citep{HsiehKnoledge,ZhangImbalanced}, clustering \citep{Baiclustering,Leeclustering} and other fields \citep{YangAnImproved,sun2011multi,xie2019multi,sun2018multiview,sun2015semisupervised,li2021machine}. 
To enhance its predictive performance and computational efficiency \citep{HuA}, the twin SVM (TSVM) has been developed by \cite{Khemchandani2007Twin}, which generates two non-parallel hyperplanes where each class is close to one and away from the other by solving a pair of smaller sized QPPs \citep{xie2018domain}. 

It is well-known that TSVM \citep{Khemchandani2007Twin} is initially designed for the binary classification problem. \hl{Since most real-life applications are related to multi-class classifications \mbox{\citep{chen2017multiple}} such as activity recognition, speaker identification and text categorization, extending it to multi-classification problems is of great significance.
Based on this, researchers have proposed many strategies, where ``one-versus-one'' (OVO), ``one-versus-rest'' (OVR) and ``one-versus-one-versus-rest'' (OVOVR) are three of the most commonly used methods \mbox{\citep{Karnik1999,XieExtending,Xu2013A,wang2021twin}}.} These different strategies have been reviewed by \cite{DingReview}. For example, \cite{Xu2013A} have proposed a multi-class classification algorithm with the OVOVR structure and produced better forecasting results than other strategies. \hl{Furthermore, some two-class techniques are often not helpful when being directly applied to the multi-class problem \mbox{\citep{zhou2005training}}, the OVOVR strategy has received much attention \mbox{\citep{pang2018scaling,PanSafe,pang2019multi}}. For example, \mbox{\cite{PanSafe}} have designed safe screening rules for accelerating the classification of TSVM. To extend it to twin multi-class SVM, \mbox{\cite{pang2018scaling}} have proposed a safe sample elimination rule for accelerating the classification. These methods can effectively accelerate the classification, whereas, efficiently obtaining the optimal regularization parameters is not involved. In this work, we aim to explore the efficient model selection \textit{w.r.t.} the regularization parameters for twin multi-class SVM based on the OVOVR strategy.}

To get the optimal regularization parameter, it is very time-consuming for the traditional grid search method, especially for multi-parameter models. 
Since \cite{Hastie2004} have proposed the entire regularization path for SVM, exploring the solution path algorithm has become one of the most efficient methods to handle the efficient model selection problem for SVM and its extensions. As can be seen in \citep{Hastie2004}, the solution path algorithm aims to establish the piecewise linear relationships between the regularization parameters and the Lagrangian multipliers. Since then, researchers have also proposed some methods to explore the solution path algorithm for TSVMs. For example, a new solution-path approach for the pinball TSVM has been proposed by \cite{YangPath} and we also have proposed a fast regularization parameter tuning algorithm for TSVM \citep{zhou2022tsvmpath}. However, the entirely regularized path algorithm for multi-class problems is much more difficult due to their intrinsic complexity. Thus, there have not been the entirely regularized solution path algorithm for twin multi-class SVM with the OVOVR structure. The reason is that the involved relationships between the regularization parameters and Lagrangian multipliers are quite complex. On the one hand, a simple yet effective initialization algorithm is hard to design. On the other hand, there are many complicated cases to consider for the OVOVR structure, so designing an efficient updating strategy is not easy. To address these problems, we propose an efficient regularized solution path algorithm for twin multi-class SVM.

In this work, twin multi-class SVM with the OVOVR strategy is transformed into two sub optimization models and the corresponding solution path algorithm is then proposed. Specifically, a novel sample data set partition strategy is first adopted, which is the basis for the model construction. Then, combining the linear equations and block matrix theory, the Lagrangian multipliers are proved to be piecewise linear \textit{w.r.t.} the regularization parameters, so that the regularization parameters are continuously updated by only solving the break points. Next, Lagrangian multipliers are proved to be 1 as the regularization parameter approaches infinity, thus, a simple yet effective initialization algorithm is devised. Finally, eight kinds of events are defined to seek for the starting event for the next iteration. Extensive experimental results on several UCI data sets show that the proposed algorithm can achieve comparable classification performance without solving any quadratic programming problem. 
Code will be available at https://github.com/ZhouKanglei/TwinMultiPath. 

The main contributions of this work are summarized as follows:

\begin{enumerate}
	\item A novel sample set partition strategy is adopted and the Lagrangian multipliers of the twin multi-class support vector machine are proved to be piecewise linear \textit{w.r.t.} the regularization parameters.
	
	\item Eight kinds of events are defined and the entire solution path algorithm is proposed, in which the regularization parameters are continuously updated by only solving the break points.
	
	\item Lagrangian multipliers are proved to be 1 as the regularization parameter approaches infinity and a simple initialization algorithm is presented, thus extending the search space of the regularization parameter to $(0, +\infty)$.
\end{enumerate}


The rest of this work is structured: \Cref{Related_work} briefly reviews the related work. \Cref{Problem} reviews the basic concept of TSVM. Details of the solution path algorithm for TSVM are introduced in \Cref{Solution}. \Cref{Experiments} shows a lot of experimental results. The conclusions of the whole paper are drawn in \Cref{Conclusion}.

\section{Related Work} \label{Related_work}
In this section, we briefly review different TSVM extensions, multi-classification strategies for TSVM and solution path algorithms, respectively.

\subsection{TSVM and Extensions}
In the last two decades, significant research achievements have been made on TSVM, including the least squares TSVM (LSTSVM), weighted TSVM (WTSVM), projection TSVM (PTSVM), \etc.

In 2009, \cite{KumarLeast} have presented LSTSVM, which introduces the concept of proximal SVM (PSVM) to the original problem of TSVM. Since only two linear equations are considered to obtain the result, the solution speed is improved a lot instead of solving two QPPs with constraints. Based on LSTSVM, many researchers have proposed different improved versions \citep{TanveerRobust, YXuXLPanZJZhou, deLimaImprovements}. To solve the semi-positive definite problem in LSTSVM, which only satisfies the empirical risk minimization, \cite{TanveerRobust} have designed a robust energy-based LSTSVM. This method uses the energy model to solve the problem of imbalanced sample data and overcomes the influence of outliers and noise. 
\cite{YXuXLPanZJZhou} have applied the prior structure information of the data to LSTSVM and constructed the structural LSTSVM. Due to the inclusion of data distribution information in the module, it has good generalization performance and short time consumption.

In 2012, based on local information, \cite{YeWeighted} have proposed a WTSVM to alleviate the problem that similar information between any two data points in the same class cannot be utilized in TSVM. 
To reduce the influence of noise, \cite{Li2016Online} have proposed a new weighting mechanism based on LSTSVM. 
\cite{XuK-nearest} has developed K-nearest neighbor (KNN)-based weighted multi-class TSVM, where the weight matrix is introduced into the objective function to explore the local information in the class, and two weight vectors are introduced into the constraint condition to find the inter-class information.

In 2011, \cite{ChenProjection} have proposed PTSVM. The idea is to find two projection directions, each one corresponds to a projection direction. The algorithm recursively generates multiple projection axes for each class, which overcomes the problem of singular values and improves the performance of the algorithm.
Furthermore, \cite{XieMulti} have presented multi-view Laplacian TSVM by combining it with semi-supervised learning. \cite{TomarA} have proposed the multi-class classification of LSTSVM by extending it to the contract state of multi-class classification. \cite{WZY2017} have developed an improved $\rho$-twin bounded SVM, which can effectively avoid the problem of matrix irreversibility in solving dual problems and has a strong generalization ability in processing large-scale data sets.

\subsection{Multi-classification Strategies}
To generalize the standard TSVM to the multi-classification problems \citep{weston1999support,crammer2002learnability}, researchers have proposed many classification strategies such as OVO, OVR, OVOVR, all-together, \etc. 

\mbox{\cite{Karnik1999}} has adopted the OVO classification strategy to establish a classifier between any two categories of samples. For the sample set with $K (K \geq 3)$ categories, $K(K - 2) / 2$ binary classification classifiers need to be constructed. And the category of the sample is determined according to its maximum vote \citep{HsuAComparison}. Obviously, the disadvantage of this classifier is that the rest samples are not considered.
\hlfour{In 2008, \mbox{\cite{cong2008efficient}} combined the OVR strategy with TSVM to achieve efficient speaker recognition.} In 2013, \cite{XieExtending} proposed a novel OVR TSVM for the multi-class classification problem and analyzed its efficiency theoretically. 
For the OVR TSVM,
by constructing $K$ binary classifiers, the $i$-th sample and the remaining samples can be distinguished by the $i$-th classifier. For unknown samples, they can be classified into the category with the maximum  confidence \textit{w.r.t.} the decision function. However, this method ignores the class imbalance problem. By combining the above strategies, the OVOVR TSVM, termed as Twin-KSVC, is proposed by \cite{Xu2013A}, which can yield better classification accuracy in comparison with other structures. \hl{Since then, this OVOVR structure has attracted much attention by researchers \mbox{\citep{XuK-nearest,pang2018scaling,pang2019multi}}. For example, \mbox{\cite{pang2018scaling}} have designed a safe sample elimination rule to identify and delete many redundant samples of all classes, so the scale of dual problems can be reduced a lot.} There are other ways \citep{DingReview, LopezANoval} to solve the multi-classification problem based on TSVM,  such as decision tree based TSVM \citep{SunBind}, directed acyclic graph based LSTSVM \citep{ZhangMulti}, ``rest-versus-one" strategy based TSVM \citep{YangMultiple}. \hl{By the way, these strategies can also be applied in SVMs for solving the multi-classification problem. It is noted that for SVMs, the all-together strategy \mbox{\citep{weston1999support}} that cannot be ignored needs to consider only one optimization problem, whereas, it is much sophisticated for practical implementations \mbox{\cite{crammer2002learnability}}.}

\hl{In addition, some studies \mbox{\citep{chen2017multiple,wang2021twin}} for multi-class classification have also achieved considerable accuracy performance. However, they have not focused on the regularization parameter tuning. In this work, we highlight the fast regularization parameter tuning with comparable prediction performance. For example, \mbox{\cite{chen2017multiple}} require solving $K$ QPPs to obtain $K$ hyperplanes for the multi-classification problem, while ours does not need to solve any QPP.}

\subsection{Solution Path Algorithms}
For the regularization parameter optimization problem, the traditional grid search method is very time-consuming \citep{PanSafe}. Recently, many fast algorithms for regularized parametric solutions \citep{Hastie2004, WangHybid2008, Gangtwo, WangA2008, OgawaSafe,  PanSafe, YangPath, OngPath, HuangPath, GuPath} have been proposed.

By fitting each cost parameter and the entire path of the solution of SVM, \cite{Hastie2004} have presented the regularized solution path algorithm of SVM. \cite{WangHybid2008} have developed the hybrid huberized SVM by using the hybrid hinge loss function and elastic network penalty, and developed an entire regularization algorithm for the hybrid huberized SVM. \cite{Gangtwo} have proposed a two-dimensional solution path for support vector regression to accelerate the process of parameter tuning. \cite{WangA2008} have designed the regression model of $\epsilon$-SVM, and proved that the solution of the model was piecewise linear \textit{w.r.t.} the parameter $\epsilon$. \cite{OgawaSafe} reduce the cost of training by introducing safe screening criteria into the parameter tuning process of SVM. In 2017, the safe screening criteria of linear TSVM and nonlinear TSVM \citep{PanSafe} are proposed to accelerate the parameter tuning process when multiple parameter models are included. In 2018, \mbox{\cite{YangPath}} have developed a new solution path approach for the pinball TSVM, where the starting point of the path could be achieved analytically without solving the optimization problem.

\section{Problem and Model} \label{Problem}
In this section, we first give the common notations with their meanings. Then, we briefly review the foundation of Twin multi-class SVM. Finally, the model transformation and the partition strategy \textit{w.r.t.} two sub-optimization problems are elaborated, respectively. 

\subsection{Notations}
Unless otherwise specified in this work, the normal bold $\mathbf{symbol}$ indicates the matrix or tensor, the italic bold $\bm{symbol}$ indicates the vector, the italic-only $symbol$ indicates the variable, and the normal $\mathrm{symbol}$ indicates the constant. Furthermore, all the sets are represented with the \textit{calligraphic} font.

Given a training data set 
$\mathcal{T} = \{(\bm{x}_{1}, y_1), (\bm{x}_{2}, y_2), \cdots, (\bm{x}_{n}, y_n)\}$, 
where $\bm{x_i} \in \mathbb{R} ^{m}$ $(i = 1, 2, \cdots, n)$ are training instances and $y_i \in \{1, 2, \cdots, K\}~(K \geq 3)$ are the corresponding labels. 
For notation convenience, we denote $\mathcal{A}$ and $\mathcal{B}$ as two different classes of samples selected from the data set $\mathcal{T}$, and the rest samples are denoted as $\mathcal{C}$. Furthermore, samples sets $\mathcal{A}$, $\mathcal{B}$ and $\mathcal{C}$ are labeled as classes ``$+1$'', ``$-1$'' and ``$0$'', respectively. 
Let 
$\mathcolorbox{myhlcolortwo}{\mathbf{A} \in \mathbb{R} ^ {n_{\mathrm{A}} \times m}} = [\bm{x}_{1}^{\top}; \bm{x}_{2}^{\top}; \cdots; \bm{x}_{n_{\mathrm{A}}}^{\top}]$, 
$\mathcolorbox{myhlcolortwo}{\mathbf{B} \in \mathbb{R} ^ {n_{\mathrm{B}} \times m}} = [\bm{x}_{1}^{\top}; \bm{x}_{2}^{\top}; \cdots; \bm{x}_{n_{\mathrm{B}}}^{\top}]$ and 
$\mathcolorbox{myhlcolortwo}{\mathbf{C} \in \mathbb{R} ^ {n_{\mathrm{C}} \times m}} = [\bm{x}_{1}^{\top}; \bm{x}_{2}^{\top}; \cdots; \bm{x}_{n_{\mathrm{C}}}^{\top}]$ stand for the sample matrices consisting of 
$\mathcal{A}$, $\mathcal{B}$ and $\mathcal{C}$ respectively, 
where $n = n_{\mathrm{A}} + n_{\mathrm{B}} + n_{\mathrm{C}}$.

\subsection{Twin Multi-class Support Vector Machine}
For the data set with $K$ classes, the twin multi-class support vector machine needs to construct $K(K-1)/2$ binary classifiers, which separate the two categories of samples $\mathcal{A}$ and $\mathcal{B}$ by seeking for two non-parallel hyperplanes 
$f_{1}:  \bm{x} ^ {\top} \bm{w}_{1} + b_{1} = 0$ and 
$f_{2}: \bm{x} ^ {\top} \bm{w}_{2} + b_{2} = 0$. 
The above problem can be solved by the following two QPPs:
\begin{equation}
\label{eqn1}
\begin{aligned}	
\min \limits_{\bm{w}_{1}, b_{1}, \bm{\xi}, \bm{\pi}} ~ 
& \frac{1}{2} || \mathbf{A} \bm{w}_{1} +  b_{1} \bm{e}_{n_{\mathrm{A}}} || ^ {2}
+ c_{1} \bm{e}_{n_{\mathrm{B}}} ^ {\top} \bm{\xi}
+ c_{2} \bm{e}_{n_{\mathrm{C}}} ^ {\top} \bm{\pi} \\
\text{s.t.} ~ 
& - (\mathbf{B} \bm{w}_{1} + b_{1} \bm{e}_{n_{\mathrm{B}}}) + \bm{\xi} \geq \bm{e}_{n_{\mathrm{B}}}, \\
& - (\mathbf{C} \bm{w}_{1} + b_{1} \bm{e}_{n_{\mathrm{C}}}) + \bm{\pi} \geq (1 - \epsilon) \bm{e}_{n_{\mathrm{C}}}, \\
& \bm{\xi} \geq 0\bm{e}_{n_{\mathrm{B}}}, ~ \bm{\pi} \geq 0 \bm{e}_{n_{\mathrm{C}}},
\end{aligned}
\end{equation}
and
\begin{equation}
\label{eqn2}
\begin{aligned}	
\min \limits_{\bm{w}_2, b_{2}, \bm{\eta}, \bm{\zeta}} ~ 
& \frac{1}{2} || \mathbf{B} \bm{w}_2 + b_{2} \bm{e}_{n_{\mathrm{B}}}|| ^ {2}
+ c_{3} \bm{e}_{n_{\mathrm{A}}} ^ {\top} \bm{\eta}
+ c_{4} \bm{e}_{n_{\mathrm{C}}} ^ {\top} \bm{\zeta} \\
\text{s.t.} ~ 
& (\mathbf{A} \bm{w}_2 + b_{2} \bm{e}_{n_{\mathrm{A}}}) + \bm{\eta} \geq \bm{e}_{n_{\mathrm{A}}}, \\
& (\mathbf{C} \bm{w}_{2} + b_{2} \bm{e}_{n_{\mathrm{C}}}) + \bm{\zeta} \geq (1 - \epsilon) \bm{e}_{n_{\mathrm{C}}}, \\
& \bm{\eta} \geq 0 \bm{e}_{n_{\mathrm{A}}}, ~ \bm{\zeta} \geq 0 \bm{e}_{n_{\mathrm{C}}},
\end{aligned}
\end{equation}
\hl{where 
$\bm{e}_{n_{\mathrm{A}}} \in \mathbb{R} ^ {n_{\mathrm{A}}}$, 
$\bm{e}_{n_{\mathrm{B}}} \in \mathbb{R} ^ {n_{\mathrm{B}}}$
and 
$\bm{e}_{n_{\mathrm{C}}} \in \mathbb{R} ^ {n_{\mathrm{C}}}$ are three unit vectors, and
$\bm{w}_{1} \in \mathbb{R} ^ {m}$ and
$\bm{w}_2 \in \mathbb{R} ^ {m}$ are coefficient vectors,  
$b_{1} \in \mathbb{R}$ and
$b_{2}\in \mathbb{R}$ are bias parameters, and 
$\bm{\xi} \in \mathbb{R} ^ {n_{\mathrm{B}}}$, 
$\bm{\pi} \in \mathbb{R} ^ {n_{\mathrm{C}}}$,
$\bm{\eta} \in \mathbb{R} ^ {n_{\mathrm{A}}}$ and
$\bm{\zeta} \in \mathbb{R} ^ {n_{\mathrm{C}}}$ are vectors of slack variables.}

\begin{figure}
	\centering  
	\includegraphics[width=\linewidth]{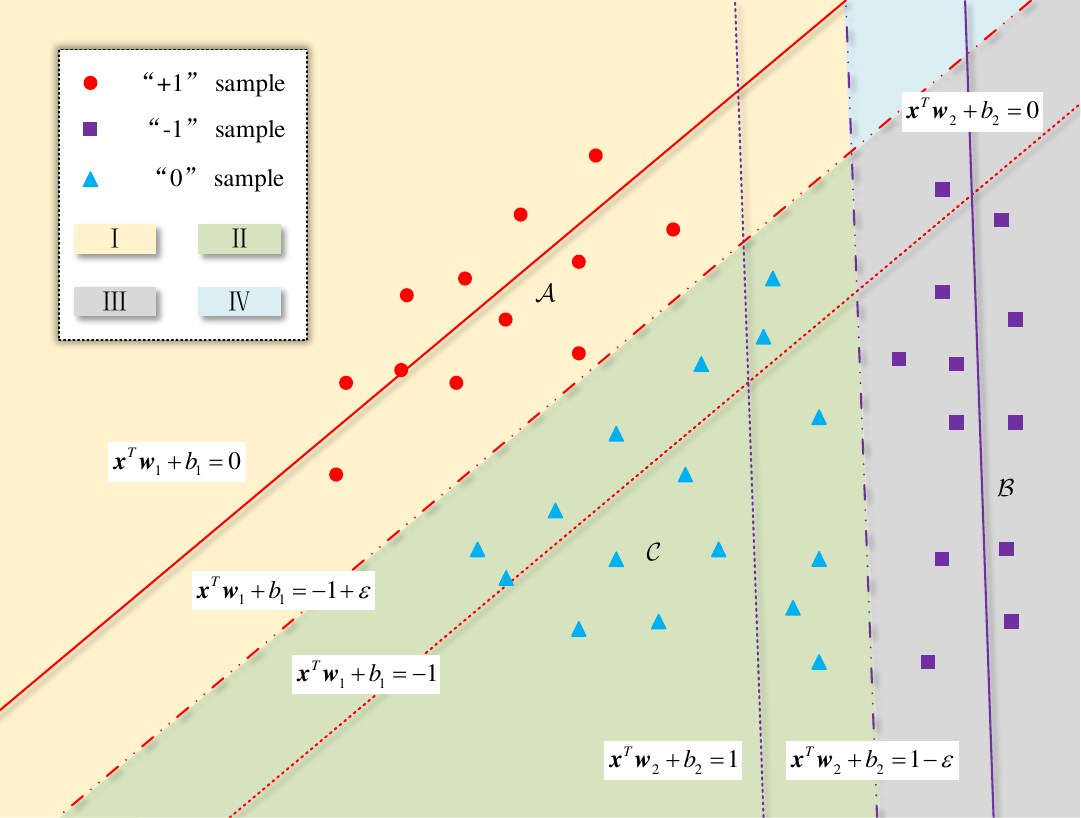}
	\caption{Two-dimensional illustration of one combination $(K_i, K_j)$ for twin multi-class support vector machine: The solid red line and the purple one represent two hyperplanes respectively; the red circle and the purple square represent two classes of samples, labeled as ``$+1$'' and ``$-1$'' respectively; the blue triangle belongs to the remaining classes, labeled as ``0''. The region can be divided into four areas, \ie, Region \uppercase\expandafter{\romannumeral1} to \uppercase\expandafter{\romannumeral4} respectively.}
	\label{Fig_0}
\end{figure}

The two-dimensional illustration of one combination $(K_i, K_j)$ for twin multi-class
support vector machine is schematically depicted in \Cref{Fig_0}. Both classes of samples are as close to the corresponding hyperplane as possible and away from the other, and the remaining samples $\mathcal{C}$ are mapped to the Region \uppercase\expandafter{\romannumeral4} in \Cref{Fig_0} between the two nonparallel hyperplanes.

\subsection{Model Transformation of the QPP \eqref{eqn1}}
Let $c_{1} = c_{2} = \frac{1}{\lambda_1}$, then the QPP (\ref{eqn1}) can be simplified as
\begin{equation}
\label{eqn3}	
\begin{aligned}
\min \limits_{\bm{w}_{1}, b_{1}, \bm{\xi}, \bm{\pi}} ~ 
& \frac{\lambda_1}{2} || \mathbf{A} \bm{w}_{1} +  b_{1} \bm{e}_{n_{\mathrm{A}}} || ^ {2}
+ \bm{e}_{n_{\mathrm{B}}} ^ {\top} \bm{\xi}
+ \bm{e}_{n_{\mathrm{C}}} ^ {\top} \bm{\pi} \\
\text{s.t.} ~ 
& - (\mathbf{B} \bm{w}_{1} + b_{1} \bm{e}_{n_{\mathrm{B}}}) + \bm{\xi} \geq \bm{e}_{n_{\mathrm{B}}}, \\
& - (\mathbf{C} \bm{w}_{1} + b_{1} \bm{e}_{n_{\mathrm{C}}}) + \bm{\pi} \geq (1 - \epsilon) \bm{e}_{n_{\mathrm{C}}}, \\
& \bm{\xi} \geq 0\bm{e}_{n_{\mathrm{B}}}, ~ \bm{\pi} \geq 0 \bm{e}_{n_{\mathrm{C}}}.
\end{aligned}
\end{equation}
The Lagrangian function of the QPP (\ref{eqn3}) can be constructed as
\begin{equation}\label{eqn4}
\begin{aligned}
\mathcal{L}_1 = 
& \frac{\lambda_1}{2} || \mathbf{A} \bm{w}_{1} +  b_{1} \bm{e}_{n_{\mathrm{A}}} || ^ {2}
+ \bm{e}_{n_{\mathrm{B}}} ^ {\top} \bm{\xi}
+ \bm{e}_{n_{\mathrm{C}}} ^ {\top} \bm{\pi} \\
& - {\bm{\alpha}^{\top}} [- (\mathbf{B} \bm{w}_{1} + b_{1} \bm{e}_{n_{\mathrm{B}}}) + \bm{\xi} - \bm{e}_{n_{\mathrm{B}}}] \\
& - {\bm{\beta}^{\top}} [- (\mathbf{C} \bm{w}_{1} + b_{1} \bm{e}_{n_{\mathrm{C}}}) + \bm{\pi} - (1 - \epsilon) \bm{e}_{n_{\mathrm{C}}}] \\
& - {\bm{\gamma}^{\top}} \bm{\xi} - {\bm{\omega}^{\top}} \bm{\pi},
\end{aligned}
\end{equation}
where 
$\bm{\alpha} \in \mathbb{R}^{n_{\mathrm{B}}}$, 
$\bm{\beta} \in \mathbb{R}^{n_{\mathrm{C}}}$, 
$\bm{\gamma} \in \mathbb{R}^{n_{\mathrm{B}}}$ and
$\bm{\omega} \in \mathbb{R}^{n_{\mathrm{C}}}$ 
are the \hl{non-negative} Lagrangian multipliers vectors. 
Set the partial derivatives of \Cref{eqn4} \textit{w.r.t.} $\bm{w}_{1}$, $b_{1}$, $\bm{\xi}$ and $\bm{\pi}$ to 0, we can obtain
\begin{align}
\frac {\partial \mathcal{L}_1} {\partial \bm{w}_{1}}
=& \lambda_1 {\mathbf{A} ^ {\top}}(\mathbf{A} \bm{w}_{1} + b_{1} \bm{e}_{n_{\mathrm{A}}})
+ \mathbf{B}^{\top} \bm{\alpha}
+ \mathbf{C}^{\top} \bm{\beta} = 0 \bm{e}_m,  \label{eqn5} \\
\frac {\partial \mathcal{L}_1} {\partial b_{1}}
=& \lambda_1 {\bm{e}_{n_{\mathrm{A}}} ^ {\top}} (\mathbf{A} \bm{w}_{1} 
+ b_{1} \bm{e}_{n_{\mathrm{A}}}) 
+ \bm{e}_{n_{\mathrm{B}}}^{\top} \bm{\alpha}
+ \bm{e}_{n_{\mathrm{C}}}^{\top} \bm{\beta} = 0, \label{eqn6} \\
\frac {\partial \mathcal{L}_1} {\partial \bm{\xi}}
=& \bm{e}_{n_{\mathrm{B}}} - \bm{\alpha} - \bm{\gamma} = 0 \bm{e}_{n_{\mathrm{B}}}, \label{eqn7} \\
\frac {\partial \mathcal{L}_1} {\partial \bm{\pi}}
=& \bm{e}_{n_{\mathrm{C}}} - \bm{\beta} - \bm{\omega} = 0 \bm{e}_{n_{\mathrm{C}}}. \label{eqn8} 
\end{align} 
According to the Karush-Kuhn-Tucker (KKT) conditions, we have
\begin{align}
{\bm{\alpha}^{\top}} [- (\mathbf{B} \bm{w}_{1} + b_{1} \bm{e}_{n_{\mathrm{B}}}) + \bm{\xi} - \bm{e}_{n_{\mathrm{B}}}] = 0, \label{eq_KKT_1_1} \\
{\bm{\beta}^{\top}} [- (\mathbf{C} \bm{w}_{1} + b_{1} \bm{e}_{n_{\mathrm{C}}}) + \bm{\pi} - (1 - \epsilon) \bm{e}_{n_{\mathrm{C}}}] = 0, \label{eq_KKT_1_2} \\
{\bm{\gamma}^{\top}} \bm{\xi} = 0, \label{eq_KKT_1_3} \\
{\bm{\omega}^{\top}} \bm{\pi} = 0. \label{eq_KKT_1_4} 
\end{align}
Combining \Cref{eqn5,eqn6}, we can obtain
\begin{equation} \label{eq_bmatrix_1}
	\lambda_1 
	\begin{bmatrix}
		\mathbf{A}^{\top} \\
		\bm{e}_{n_{\mathrm{A}}}^{\top}
	\end{bmatrix}
	\begin{bmatrix}
		\mathbf{A} &
		\bm{e}_{n_{\mathrm{A}}}
	\end{bmatrix}
	\begin{bmatrix}
		\bm{w}_1 \\
		b_1
	\end{bmatrix}
	+
	\begin{bmatrix}
		\mathbf{B}^{\top} \\
		\bm{e}_{n_{\mathrm{B}}}^{\top}
	\end{bmatrix}
	\bm{\alpha}
	+
	\begin{bmatrix}
		\mathbf{C}^{\top} \\
		\bm{e}_{n_{\mathrm{C}}}^{\top}
	\end{bmatrix}
	\bm{\beta}
	=
	0 \bm{e}_{m + 1}.
\end{equation}
Let 
$\mathbf{F} = [\mathbf{A}, \bm{e}_{n_{\mathrm{A}}}]$,
$\mathbf{G} = [\mathbf{B}, \bm{e}_{n_{\mathrm{B}}}]$,
$\mathbf{H} = [\mathbf{C}, \bm{e}_{n_{\mathrm{C}}}]$ and
$\bm{u} = [\bm{w}_1; b_1]$. 
The above equation can be rewritten as
\begin{equation}
	\lambda_1 \mathbf{F}^{\top} \mathbf{F} \bm{u} 
	+
	\mathbf{G}^{\top} \bm{\alpha}
	+
	\mathbf{H}^{\top} \bm{\beta}
	=
	0 \bm{e}_{m + 1}.
\end{equation}
From \Cref{eq_bmatrix_1}, the solution of the QPP \eqref{eqn3} can be obtained 
when the matrix $\mathbf{F}^{\top} \mathbf{F}$ is invertible.
\begin{equation}
\bm{u} 
= 
- \frac{1}{\lambda_1} 
\mathbf{F}_{\mathrm{inv}}
(\mathbf{G}^{\top} \bm{\alpha}
+
\mathbf{H}^{\top} \bm{\beta}), \label{eqn13}
\end{equation}
where $\mathbf{F}_{\mathrm{inv}}$ represents the inverse matrix of $\mathbf{F} ^ {\top} \mathbf{F}$. Since the matrix $\mathbf{F} ^ {\top} \mathbf{F}$ is always semi-positive definite, \hlthree{we add the regularization term $\delta \mathbf{I}$ to avoid the ill-conditioning case in this work, where $\delta$ is a small positive real number and $\mathbf{I}$ is an identity matrix in $\mathbb{R} ^ {(m + 1) \times (m + 1)}$. Thus,}
\begin{equation}
\mathbf{F}_{\mathrm{inv}}
=  
(\mathbf{F} ^ {\top} \mathbf{F} + \delta \mathbf{I}) ^ {-1}. \label{eq_F_inv} 
\end{equation}

According to \Cref{eqn13}, the function $f_1$ can be represented as
\begin{equation}
	f_1(\bm{x}) = 
	- \frac{1}{\lambda_1} 
	\begin{bmatrix}
		\bm{x}^{\top} &  1
	\end{bmatrix} 
	\mathbf{F}_{\mathrm{inv}}
	(\mathbf{G}^{\top} \bm{\alpha}
	+
	\mathbf{H}^{\top} \bm{\beta}). \label{eq_f_1}\\
\end{equation}

\subsection{Partition Strategies of the QPP \eqref{eqn3}} \label{sec_partition_strategy}
From \Cref{eqn13}, the Lagrangian multipliers $\bm{\alpha}$ and $\bm{\beta}$ of the QPP \eqref{eqn3} correspond to the sample in $\mathcal{B}$ and $\mathcal{C}$, but not to that in $\mathcal{A}$. 
Hence, the samples in $\mathcal{B}$ and $\mathcal{C}$ need to be divided.

\subsubsection{Partition Strategy of Samples in $\mathcal{B}$} \label{sec_partition_B_1}
\hl{Combining the non-negative properties of Lagrangian multipliers $\bm{\alpha}, \bm{\gamma}$ with \mbox{\Cref{eqn7}}, it is easy to obtain $0\bm{e}_{n_\mathrm{B}} \leq \bm{\alpha}, \bm{\gamma} \leq \bm{e}_{n_\mathrm{B}}$.}
According to \cite{Hastie2004}, by combining the constraint conditions in \Cref{eqn5,eqn6,eqn7,eqn8} and the KKT conditions in \Cref{eq_KKT_1_1,eq_KKT_1_2,eq_KKT_1_3,eq_KKT_1_4} of the QPP \mbox{\eqref{eqn3}}, \hltwo{we see that the sample $\bm{x}_i~(i \in \mathcal{B})$ can be discussed in three situations: 
when $-(\bm{x}_{i}^{\top} \bm{w}_{1} + b_{1}) < 1$, $\alpha_i = 1$; 
when $-(\bm{x}_{i}^{\top} \bm{w}_{1} + b_{1}) > 1$, $\alpha_i = 0$; 
when $-(\bm{x}_{i}^{\top} \bm{w}_{1} + b_{1}) = 1$, $\alpha_i$ can lie between 0 and 1.}

\begin{figure}[ht]
	\centering
	\includegraphics[width=\linewidth]{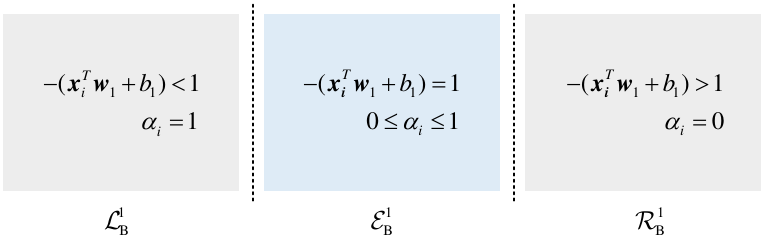}	
	\caption{Partition of the sample set $\mathcal{B}$ for the QPP \eqref{eqn3}.}
	\label{fign1}
\end{figure}

Therefore, the samples in $\mathcal{B}$ can be divided into three sets, \ie,
$\mathcal{L}_{\mathrm{B}} ^ {1} = \{ i | -(\bm{x}_i^{\top} \bm{w}_{1} + b_{1}) < 1 \}$,
$\mathcal{E}_{\mathrm{B}} ^ {1} = \{ i | -(\bm{x}_i^{\top} \bm{w}_{1} + b_{1}) = 1 \}$, and
$\mathcal{R}_{\mathrm{B}} ^ {1} = \{ i | -(\bm{x}_i^{\top} \bm{w}_{1} + b_{1}) > 1 \}$, 
as shown in \Cref{fign1}.

\subsubsection{Partition Strategy for Samples in $\mathcal{C}$}
Similar to \Cref{sec_partition_B_1}, 
\hltwo{it is easy to see that the sample $\bm{x}_k~(k \in \mathcal{C})$ can be discussed in three situations: 
when $-(\bm{x}_{k}^{\top} \bm{w}_{1} + b_{1}) < 1 - \epsilon$, $\beta_k = 1$; 
when $-(\bm{x}_{k}^{\top} \bm{w}_{1} + b_{1}) > 1 - \epsilon$, $\beta_k = 0$; 
when $-(\bm{x}_{k}^{\top} \bm{w}_{1} + b_{1}) = 1 - \epsilon$, $\beta_k$ can lie between 0 and 1.}

\begin{figure}[ht]
	\centering 
	\includegraphics[width=\linewidth]{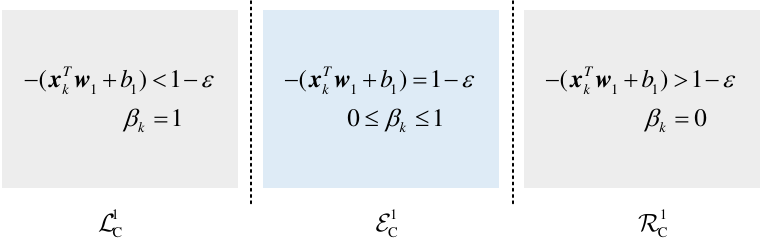}			
	\caption{Partition of the sample set $\mathcal{C}$ for the QPP \eqref{eqn3}.} 
	\label{fign2}
\end{figure}

Therefore, the samples in $\mathcal{C}$ can be divided into three sets, \ie,
$\mathcal{L}_{\mathrm{C}} ^ {1} = \{ k | -(\bm{x}_{k}^{\top} \bm{w}_{1}  + b_{1}) < 1 - \epsilon \}$,
$\mathcal{E}_{\mathrm{C}} ^ {1} = \{ k | -(\bm{x}_{k}^{\top} \bm{w}_{1}  + b_{1}) = 1 - \epsilon \}$, and
$\mathcal{R}_{\mathrm{C}} ^ {1} = \{ k | -(\bm{x}_{k}^{\top} \bm{w}_{1}  + b_{1}) > 1 - \epsilon \}$, 
as shown in \Cref{fign2}.

\subsection{Model Transformation of the QPP \eqref{eqn2}}
Let $c_{3} = c_{4} = \frac{1}{\lambda_2}$, the QPP \eqref{eqn2} can be rewritten as
\begin{equation}
\label{eqn2_t}
\begin{aligned}	
\min \limits_{\bm{w}_2, b_{2}, \bm{\eta}, \bm{\zeta}} ~ 
& \frac{\lambda_2}{2} || \mathbf{B} \bm{w}_2 + b_{2} \bm{e}_{n_{\mathrm{B}}}|| ^ {2}
+ \bm{e}_{n_{\mathrm{A}}} ^ {\top} \bm{\eta}
+ \bm{e}_{n_{\mathrm{C}}} ^ {\top} \bm{\zeta} \\
\text{s.t.} ~ 
& (\mathbf{A} \bm{w}_2 + b_{2} \bm{e}_{n_{\mathrm{A}}}) + \bm{\eta} \geq \bm{e}_{n_{\mathrm{A}}}, \\
& (\mathbf{C} \bm{w}_{2} + b_{2} \bm{e}_{n_{\mathrm{C}}}) + \bm{\zeta} \geq (1 - \epsilon) \bm{e}_{n_{\mathrm{C}}}, \\
& \bm{\eta} \geq 0 \bm{e}_{n_{\mathrm{A}}}, ~ \bm{\zeta} \geq 0 \bm{e}_{n_{\mathrm{C}}}.
\end{aligned}
\end{equation}

By constructing the Lagrangian function of \Cref{eqn2_t}, we can obtain
\begin{equation} \label{eq_L_2}
	\begin{aligned}
		\mathcal{L}_2 
		= &\frac{\lambda_2}{2} || \mathbf{B} \bm{w}_2 + b_{2} \bm{e}_{n_{\mathrm{B}}}|| ^ {2}
		 + \bm{e}_{n_{\mathrm{A}}} ^ {\top} \bm{\eta}
		 + \bm{e}_{n_{\mathrm{C}}} ^ {\top} \bm{\zeta} \\
		&- \bm{\mu}^{\top} [(\mathbf{A} \bm{w}_2 + b_{2} \bm{e}_{n_{\mathrm{A}}}) + \bm{\eta} - \bm{e}_{n_{\mathrm{A}}}] \\
		&- \bm{\rho}^{\top} [(\mathbf{C} \bm{w}_{2} + b_{2} \bm{e}_{n_{\mathrm{C}}}) + \bm{\zeta} - (1 - \epsilon) \bm{e}_{n_{\mathrm{C}}}] \\
		&- \bm{\phi}^{\top} \bm{\eta} - \bm{\psi}^{\top} \bm{\zeta},
	\end{aligned}
\end{equation}
where 
$\bm{\mu} \in \mathbb{R}^{n_{\mathrm{A}}}$, 
$\bm{\rho} \in \mathbb{R}^{n_{\mathrm{C}}}$, 
$\bm{\phi} \in \mathbb{R}^{n_{\mathrm{A}}}$ and
$\bm{\psi} \in \mathbb{R}^{n_{\mathrm{C}}}$ 
are the \hl{non-negative} Lagrangian multipliers vectors. Let the partial derivatives of Lagrangian function \eqref{eq_L_2} \textit{w.r.t.} $\bm{w}_2$, $ b_2$, $\bm{\eta}$  and $\bm{\zeta}$ be 0 respectively, and we can get
\begin{align}
\frac {\partial \mathcal{L}_2} {\partial \bm{w}_{1}}
=& \lambda_2 {\mathbf{B} ^ {\top}}(\mathbf{B} \bm{w}_{2} + b_{2} \bm{e}_{n_{\mathrm{B}}})
- \mathbf{A}^{\top} \bm{\mu}
- \mathbf{C}^{\top} \bm{\rho} = 0 \bm{e}_m,  \label{eq_partial_2_1} \\
\frac {\partial \mathcal{L}_2} {\partial b_{2}}
=& \lambda_2 {\bm{e}_{n_{\mathrm{B}}} ^ {\top}} (\mathbf{B} \bm{w}_{2} 
+ b_{2} \bm{e}_{n_{\mathrm{B}}}) 
- \bm{e}_{n_{\mathrm{A}}}^{\top} \bm{\mu}
- \bm{e}_{n_{\mathrm{C}}}^{\top} \bm{\rho} = 0, \label{eq_partial_2_2} \\
\frac {\partial \mathcal{L}_2} {\partial \bm{\eta}}
=& \bm{e}_{n_{\mathrm{A}}} - \bm{\mu} - \bm{\phi} = 0 \bm{e}_{n_{\mathrm{A}}}, \label{eq_partial_2_3} \\
\frac {\partial \mathcal{L}_2} {\partial \bm{\zeta}}
=& \bm{e}_{n_{\mathrm{C}}} - \bm{\rho} - \bm{\psi} = 0 \bm{e}_{n_{\mathrm{C}}}. \label{eq_partial_2_4} 
\end{align} 
Combining the KKT conditions with the QPP \eqref{eqn2_t}, we can obtain 
\begin{align}
	\bm{\mu}^{\top} [(\mathbf{A} \bm{w}_2 + b_{2} \bm{e}_{n_{\mathrm{A}}}) + \bm{\eta} - \bm{e}_{n_{\mathrm{A}}}] = 0, \label{eq_KKT_2_1}\\
	\bm{\rho}^{\top} [(\mathbf{C} \bm{w}_{2} + b_{2} \bm{e}_{n_{\mathrm{C}}}) + \bm{\zeta} - (1 - \epsilon) \bm{e}_{n_{\mathrm{C}}}] = 0, \label{eq_KKT_2_2} \\
	\bm{\phi}^{\top} \bm{\eta} = 0, \label{eq_KKT_2_3} \\
	\bm{\psi}^{\top} \bm{\zeta} = 0. \label{eq_KKT_2_4}
\end{align}
From \Cref{eq_partial_2_1} and \Cref{eq_partial_2_2}, we can obtain
\begin{equation} \label{eq_bmatrix_2}
	\lambda_2 
	\begin{bmatrix}
		\mathbf{B}^{\top} \\
		\bm{e}_{n_{\mathrm{B}}}^{\top}
	\end{bmatrix}
	\begin{bmatrix}
		\mathbf{B} &
		\bm{e}_{n_{\mathrm{B}}}
	\end{bmatrix}
	\begin{bmatrix}
		\bm{w}_2 \\
		b_2
	\end{bmatrix}
	-
	\begin{bmatrix}
		\mathbf{A}^{\top} \\
		\bm{e}_{n_{\mathrm{A}}}^{\top}
	\end{bmatrix}
	\bm{\mu}
	-
	\begin{bmatrix}
		\mathbf{C}^{\top} \\
		\bm{e}_{n_{\mathrm{C}}}^{\top}
	\end{bmatrix}
	\bm{\rho}
	=
	0 \bm{e}_{m + 1}.
\end{equation}
Let $\bm{v} = [\bm{w}_2; b_2]$, the above equation can be rewritten as
\begin{equation}
	\lambda_2 \mathbf{G}^{\top} \mathbf{G} \bm{v} - \mathbf{F}^{\top} \bm{\mu} - \mathbf{H}^{\top} \bm{\rho} = 0 \bm{e}_{m + 1}.
\end{equation}
Therefore, the solution of the QPP \eqref{eqn2_t} can be obtained when the matrix $\mathbf{G}^{\top} \mathbf{G}$ is invertible.
\begin{equation}
	\bm{v} = \frac{1}{\lambda_2} \mathbf{G}_{\mathrm{inv}} (\mathbf{F}^{\top} \bm{\mu} + \mathbf{H}^{\top} \bm{\rho}), \label{eq_v}
\end{equation}
\hltwo{where $\mathbf{G}_{\mathrm{inv}}$ represents the inverse matrix of $\mathbf{G} ^ {\top} \mathbf{G}$. Similar to \mbox{\Cref{eq_F_inv}}, we also add the regularization term $\delta \mathbf{I}$ to avoid the ill-conditioning case in this work.}
\begin{equation}
	\mathbf{G}_{\mathrm{inv}} = (\mathbf{G}^{\top} \mathbf{G} + \delta \mathbf{I})^{-1}. \label{eq_G_inv}
\end{equation}

According to \Cref{eq_v}, the function $f_2$ can be represented as
\begin{equation}
	f_2(\bm{x}) = 
	\frac{1}{\lambda_2} 
	\begin{bmatrix}
		\bm{x}^{\top} &  1
	\end{bmatrix} 
	\mathbf{G}_{\mathrm{inv}}
	(\mathbf{F}^{\top} \bm{\mu}
	+
	\mathbf{H}^{\top} \bm{\rho}). \label{eq_f_2}\\
\end{equation}

\subsection{Partition Strategies of the QPP \eqref{eqn2_t}}

According to \Cref{eq_v}, the Lagrangian multipliers $\bm{\mu}$ and $\bm{\rho}$ of the QPP \eqref{eqn2_t} correspond to the sample in $\mathcal{A}$ and $\mathcal{C}$, but not to that in $\mathcal{B}$. \hl{Hence, the samples in $\mathcal{A}$ and $\mathcal{C}$ need to be divided}.

\subsubsection{Partition Strategy of Samples in $\mathcal{A}$} \label{sec_partition_A}
\hl{Combining the non-negative properties of Lagrangian multipliers $\bm{\mu}, \bm{\phi}$ and \mbox{\Cref{eq_partial_2_3}}, it is easy to obtain $0\bm{e}_{n_\mathrm{A}} \leq \bm{\mu}, \bm{\phi} \leq \bm{e}_{n_\mathrm{A}}$.}
According to \cite{Hastie2004}, by combining the constraint conditions in \Cref{eq_partial_2_1,eq_partial_2_2,eq_partial_2_3,eq_partial_2_4} and the KKT conditions in \Cref{eq_KKT_2_1,eq_KKT_2_2,eq_KKT_2_3,eq_KKT_2_4} of the QPP \mbox{\eqref{eqn2_t}}, \hltwo{we see that the sample $\bm{x}_i~(i \in \mathcal{A})$ can be discussed in three situations: 
when $\bm{x}_{i}^{\top} \bm{w}_{2} + b_{2} < 1$, $\mu_i = 1$; 
when $\bm{x}_{i}^{\top} \bm{w}_{2} + b_{2} > 1$, $\mu_i = 0$; 
when $\bm{x}_{i}^{\top} \bm{w}_{2} + b_{2} = 1$, $\mu_i$ can lie between 0 and 1.}

\begin{figure}[ht]
	\centering
	\includegraphics[width=\linewidth]{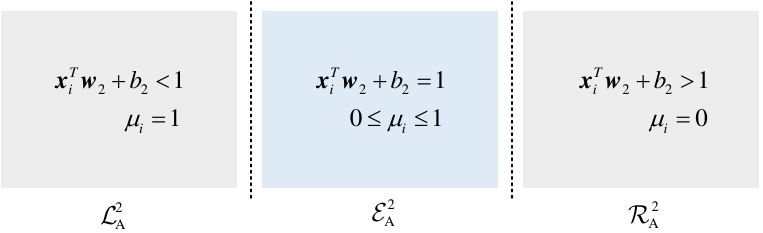}			
	\caption{Partition of the sample set $\mathcal{A}$ for the QPP \eqref{eqn2_t}.}
	\label{fign3}
\end{figure}

Therefore, the samples in $\mathcal{A}$ can be divided into three sets, \ie,
$\mathcolorbox{myhlcolor}{\mathcal{L}_{\mathrm{A}}^{2}} = \{ i | -(\bm{x}_i^{\top} \bm{w}_{2} + b_{2}) < 1 \}$,
$\mathcolorbox{myhlcolor}{\mathcal{E}_{\mathrm{A}}^{2}} = \{ i | -(\bm{x}_i^{\top} \bm{w}_{2} + b_{2}) = 1 \}$, and
$\mathcolorbox{myhlcolor}{\mathcal{R}_{\mathrm{A}}^{2}} = \{ i | -(\bm{x}_i^{\top} \bm{w}_{2} + b_{2}) > 1 \}$, 
as shown in \Cref{fign3}.

\subsubsection{Partition Strategy for Samples in $\mathcal{C}$}
Similar to \Cref{sec_partition_A}, 
\hltwo{it is easy to see that the sample $\bm{x}_k~(k \in \mathcal{C})$ can be discussed in three situations: 
when $\bm{x}_{k}^{\top} \bm{w}_{2} + b_{2} < 1 - \epsilon$, $\rho_k = 1$; 
when $\bm{x}_{k}^{\top} \bm{w}_{2} + b_{2} > 1 - \epsilon$, $\rho_k = 0$; 
when $\bm{x}_{k}^{\top} \bm{w}_{2} + b_{2} = 1 - \epsilon$, $\rho_k$ can lie between 0 and 1.}

\begin{figure}[ht]
	\centering 
	\includegraphics[width=\linewidth]{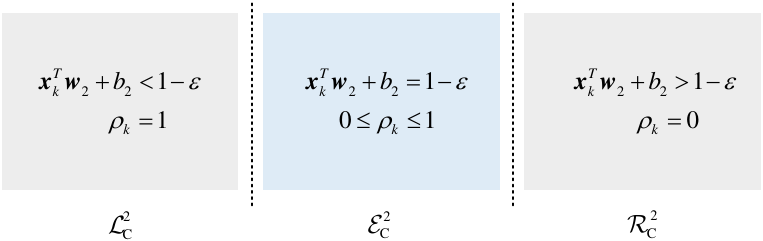}			
	\caption{Partition of the sample set $\mathcal{C}$ for the QPP \eqref{eqn2_t}.} 
	\label{fign4}
\end{figure}

Therefore, the samples in $\mathcal{C}$ can be divided into three sets, \ie,
$\mathcolorbox{myhlcolor}{\mathcal{L}_{\mathrm{C}}^{2}} = \{ k | -(\bm{x}_{k}^{\top} \bm{w}_{2}  + b_{2}) < 1 - \epsilon \}$,
$\mathcolorbox{myhlcolor}{\mathcal{E}_{\mathrm{C}}^{2}} = \{ k | -(\bm{x}_{k}^{\top} \bm{w}_{2}  + b_{2}) = 1 - \epsilon \}$, and
$\mathcolorbox{myhlcolor}{\mathcal{R}_{\mathrm{C}}^{2}} = \{ k | -(\bm{x}_{k}^{\top} \bm{w}_{2}  + b_{2}) > 1 - \epsilon \}$,
as shown in \Cref{fign4}.

\section{Solution Path for Twin Multi-class Support Vector Machine} \label{Solution}

This section aims to find the entire regularized solutions for all the values of the regularization parameter $\lambda > 0$. To this end, we first prove that the solution is piecewise linear \textit{w.r.t.} $\lambda$, which greatly reduces computational cost. Let $\lambda$ go from very large to 0, and then we just need to find all the break points for the $\lambda$ to figure out the entire solution path. 

\subsection{Piecewise Linear Theory}
\hl{
The matrices $\mathbf{F}_{\mathrm{E}}$, $\mathbf{G}_{\mathrm{E}}$ and $\mathbf{H}_{\mathrm{E}}$ are used to represent the sample matrix composed of the elbow 
$\mathcal{E}_{\mathrm{A}}$, $\mathcal{E}_B$ and $\mathcal{E}_C$. 
Let 
$n_{\mathrm{A}} = |\mathcal{E}_{\mathrm{A}}|$,
$n_{\mathrm{B}} = |\mathcal{E}_B|$,
$n_{\mathrm{C}} = |\mathcal{E}_C|$,
$n_{\mathrm{BC}} = |\mathcal{E}_B| + |\mathcal{E}_C|$ and
$n_{\mathrm{AC}} = |\mathcal{E}_{\mathrm{A}}| + |\mathcal{E}_C|$.
For convenient notations, the $l$-th step parameters are assigned a \textit{superscript} `$l$' and that of the two QPPs are assigned a \textit{superscript} `$1$' and `$2$',  respectively. For example, $\mathcal{E}_{\mathrm{B}}^{1,l}$ denotes the $l$-step index set \textit{w.r.t.} the first QPP.}

\subsubsection{Piecewise Linear \textit{w.r.t.} $\lambda_1$}

\Cref{theorem1} can be used to prove that the Lagrangian multipliers $\bm{\alpha}$ and $\bm{\beta}$ are piecewise linear \textit{w.r.t.} the regularization parameter $\lambda_1$.

\begin{theorem} \label{theorem1}
	For the QPP (\ref{eqn3}), $\lambda_1 ^ {l} > \lambda_1 > \lambda_1 ^ {l + 1}$, if the matrix
	\begin{equation} \label{eq_E_BC}
		\mathbf{E}_{\mathrm{BC}}^l = 
		\begin{bmatrix}
			\mathbf{G}_{\mathrm{E}}^l \\ 
			\mathbf{H}_{\mathrm{E}}^l
		\end{bmatrix}
		\mathbf{F}_{\mathrm{inv}}
		\begin{bmatrix}
			(\mathbf{G}_{\mathrm{E}}^l)^{\top} & (\mathbf{H}_{\mathrm{E}}^l)^{\top}
		\end{bmatrix}
	\end{equation}
	is invertible,
	then the Lagrangian multipliers 
	$\alpha_{i} ~ (i = 1, 2, \cdots, n_{\mathrm{B}}^{l})$ and $\beta_{k} ~ (k = 1, 2, \cdots, n_{\mathrm{C}}^{l})$ are piecewise linear \textit{w.r.t.} the regularization parameter $\lambda_1$ respectively,
	which can be mathematically described below:
	\begin{align}
	\alpha_{i} &= \alpha_{i} ^ {l} - (\lambda ^ {l} - \lambda ) \theta_{i} ^ {l}, \label{eqn18} \\
	\beta_{k}  &= \beta_{k} ^ {l} - (\lambda ^ {l} - \lambda ) \theta_{n_\mathrm{B}^l + k} ^ {l}, \label{eqn19}
	\end{align}
	where 
	\begin{equation} \label{eq_theta_l}
		\begin{aligned}
			\bm{\theta}^l &= (\mathbf{E}_{\mathrm{BC}}^l)^{-1} 
			\begin{bmatrix}
			 	\bm{e}_{n_{\mathrm{B}}^l} \\
			 	(1 - \epsilon) \bm{e}_{n_{\mathrm{C}}^l}
			\end{bmatrix} \\
			&= (\mathbf{E}_{\mathrm{BC}}^l)^{-1} \mathcolorbox{myhlcolor}{\bm{e}_{n_\mathrm{BC}}^l}.
		\end{aligned}
	\end{equation}
\end{theorem}

\begin{proof}
	\Cref{theorem1} can be proved in two steps, \ie, proving that Lagrangian multipliers $\alpha_{i} ~ (i = 1, 2, \cdots, n_{\mathrm{B}}^{l})$ and $\beta_{k} ~ (k = 1, 2, \cdots, n_{\mathrm{C}}^{l})$ are piecewise linear \textit{w.r.t.} the regularization parameter $\lambda_1$ respectively.
	
	First of all, the $l$-th step function of \Cref{eq_f_1} is
	\begin{equation}
		f_1^l(\bm{x}) = 
		- \frac{1}{\lambda_1^l} 
		\begin{bmatrix}
			\bm{x}^{\top} &  1
		\end{bmatrix} 
		\mathbf{F}_{\mathrm{inv}}
		(\mathbf{G}^{\top} \bm{\alpha}^l
		+
		\mathbf{H}^{\top} \bm{\beta}^l). \label{eq_f_1^l}\\
	\end{equation}

	By the following transformation about $f_1$ and $f_1^l$, we can obtain
	\begin{equation} \label{eq_f_1^l_f_1}
		\begin{aligned}
			f_1(\bm{x}) = &\frac{\lambda_1^l}{\lambda_1}f_1^l(\bm{x}) + f_1(\bm{x})
			- \frac{\lambda_1^l}{\lambda_1}f_1^l(\bm{x}) \\
			= & \frac{1}{\lambda_1} \{ 
			\lambda_1^l f_1^l(\bm{x}) +
			\begin{bmatrix}
				\bm{x}^{\top} & 1
			\end{bmatrix}
			\mathbf{F}_{\mathrm{inv}}
			[\mathbf{G}^{\top} (\bm{\alpha}^l - \bm{\alpha}) + \mathbf{H}^{\top} (\bm{\beta}^l - \bm{\beta})]
			\}.
		\end{aligned}
	\end{equation}
	For $\forall i \in \mathcal{L}_{\mathrm{B}}^{1, l} \cup \mathcal{R}_{\mathrm{B}}^{1, l}$ and
	$\forall k \in \mathcal{L}_{\mathrm{C}}^{1, l} \cup \mathcal{R}_{\mathrm{C}}^{1, l}$, 
	there are $\alpha_i^l - \alpha_i = 0$ and $\beta_k^l - \beta_k = 0$ respectively.
	Therefore, the above equation \eqref{eq_f_1^l_f_1} can be simplified as
	\begin{equation} \label{eq_f_1_system}
		\begin{aligned}
			f_1(\bm{x}) 
			= & \frac{1}{\lambda_1} \{ 
			\lambda_1^l f_1^l(\bm{x}) +
			\begin{bmatrix}
				\bm{x}^{\top} & 1
			\end{bmatrix}
			\mathbf{F}_{\mathrm{inv}}
			[\mathbf{G}_{\mathrm{E}}^{\top} (\bm{\alpha}_{\mathrm{E}}^l - \bm{\alpha}_{\mathrm{E}}) + \mathbf{H}^{\top} (\bm{\beta}^l_{\mathrm{E}} - \bm{\beta}_{\mathrm{E}})]
			\} \\
			= & \frac{1}{\lambda_1} \{ 
			\lambda_1^l f_1^l(\bm{x}) +
			\begin{bmatrix}
				\bm{x}^{\top} & 1
			\end{bmatrix}
			\mathbf{F}_{\mathrm{inv}}
			\begin{bmatrix}
				(\mathbf{G}_{\mathrm{E}}^l)^{\top} & (\mathbf{H}_{\mathrm{E}}^l)^{\top}
			\end{bmatrix}
			\begin{bmatrix}
				\bm{\alpha}_{\mathrm{E}}^l - \bm{\alpha}_{\mathrm{E}} \\
				\bm{\beta}_{\mathrm{E}}^l - \bm{\beta}_{\mathrm{E}}
			\end{bmatrix}
			\} 
			.
		\end{aligned}
	\end{equation}
	\hl{
	For $\forall i \in \mathcal{E}_{\mathrm{B}}^{1, l}$ and
	$\forall k \in \mathcal{E}_{\mathrm{C}}^{1, l}$, there are 
	$- f_1(\bm{x}_i) = - f_1^l(\bm{x}_i) = 1$ and 
	$- f_1(\bm{x}_k) = - f_1^l(\bm{x}_k) = 1 - \epsilon$ respectively.
	Substitute these two obtained conditions to \mbox{\Cref{eq_f_1_system}} and combine it with \mbox{\Cref{eq_E_BC}}, the following system of linear equation \textit{w.r.t.} the Lagrangian multipliers $\mathbf{\alpha}_{\mathrm{E}}$ and $\mathbf{\beta}_{\mathrm{E}}$ can be deduced.
	}
	\begin{equation}
	\begin{aligned}
		\begin{bmatrix}
			\mathbf{G}_{\mathrm{E}}^l \\ 
			\mathbf{H}_{\mathrm{E}}^l
		\end{bmatrix}
		\mathbf{F}_{\mathrm{inv}}
		\begin{bmatrix}
			(\mathbf{G}_{\mathrm{E}}^l)^{\top} & (\mathbf{H}_{\mathrm{E}}^l)^{\top}
		\end{bmatrix}
		\begin{bmatrix}
			\bm{\alpha}_{\mathrm{E}}^l - \bm{\alpha}_{\mathrm{E}} \\
			\bm{\beta}_{\mathrm{E}}^l - \bm{\beta}_{\mathrm{E}}
		\end{bmatrix} 
		&= \mathbf{E}_{\mathrm{BC}}^l
		\begin{bmatrix}
			\bm{\alpha}_{\mathrm{E}}^l - \bm{\alpha}_{\mathrm{E}} \\
			\bm{\beta}_{\mathrm{E}}^l - \bm{\beta}_{\mathrm{E}}
		\end{bmatrix} \\
		&= (\lambda_1^l - \lambda_1) \mathcolorbox{myhlcolor}{\bm{e}_{n_\mathrm{BC}}^l}.
	\end{aligned}
	\end{equation}
	If the matrix $\mathbf{E}_{\mathrm{BC}}^l$ is invertible, then we can obtain
	\begin{equation}
	\begin{aligned}
		\begin{bmatrix}
			\bm{\alpha}_{\mathrm{E}}^l - \bm{\alpha}_{\mathrm{E}} \\
			\bm{\beta}_{\mathrm{E}}^l - \bm{\beta}_{\mathrm{E}}
		\end{bmatrix} 
		&= (\lambda_1^l - \lambda_1) (\mathbf{E}_{\mathrm{BC}}^l)^{-1} \mathcolorbox{myhlcolor}{\bm{e}_{n_\mathrm{BC}}^l} \\
		&= (\lambda_1^l - \lambda_1) \bm{\theta}^l.
	\end{aligned}
	\end{equation}

	In conclusion, \Cref{theorem1} is proved.
\end{proof}

According to \Cref{theorem1}, it is easy to obtain the following corollary:

\begin{corollary}
	For the QPP (\ref{eqn3}), $\lambda_1 ^ {l} > \lambda_1 > \lambda_1 ^ {l + 1}$, the function $f_1$ is piecewise linear \textit{w.r.t.} $\frac{1}{\lambda_1}$, \ie,
	\begin{equation}\label{eq_fx_gx}
	\begin{aligned} 
		f_1(\bm{x}) 
		&= \frac{1}{\lambda_1}
		[ \lambda_1^l f_1^l(\bm{x}) 
		+
		(\lambda_1^l - \lambda_1)
		g^l(\bm{x}) 
		] \\
		&= \frac{\lambda_1^l}{\lambda_1} [f_1^l(\bm{x} + g^l(\bm{x})] - g^l(\bm{x}),
	\end{aligned}
	\end{equation}
	where
	\begin{equation}
		g^l(\bm{x}) 
		= 
		\begin{bmatrix}
			\bm{x}^{\top} & 1
		\end{bmatrix}
		\mathbf{F}_{\mathrm{inv}}
		\begin{bmatrix}
			(\mathbf{G}_{\mathrm{E}}^l)^{\top} & (\mathbf{H}_{\mathrm{E}}^l)^{\top}
		\end{bmatrix}
		\bm{\theta}^l.
	\end{equation}
\end{corollary}

\subsubsection{Piecewise Linear \textit{w.r.t.} $\lambda_2$}
Analogously, \Cref{theorem2} can be used to prove that the Lagrangian multipliers $\bm{\mu}$ and $\bm{\rho}$ are piecewise linear \textit{w.r.t.} the regularization parameter $\lambda_2$.

\begin{theorem} \label{theorem2}
	For the QPP (\ref{eqn2_t}), $\lambda_2 ^ {l} > \lambda_2 > \lambda_2 ^ {l + 1}$, if the matrix
	\begin{equation} \label{eq_E_AC}
		\mathbf{E}_{\mathrm{AC}}^l = 
		\begin{bmatrix}
			\mathbf{F}_{\mathrm{E}}^l \\ 
			\mathbf{H}_{\mathrm{E}}^l
		\end{bmatrix}
		\mathbf{G}_{\mathrm{inv}}
		\begin{bmatrix}
			(\mathbf{F}_{\mathrm{E}}^l)^{\top} & (\mathbf{H}_{\mathrm{E}}^l)^{\top}
		\end{bmatrix}
	\end{equation}
	is invertible, 
	then the Lagrangian multipliers 
	$\mu_{i} ~ (i = 1, 2, \cdots, n_{\mathrm{A}}^{l})$ and $\rho_{k} ~ (k = 1, 2, \cdots, n_{\mathrm{C}}^{l})$ are piecewise linear \textit{w.r.t.} the regularization parameter $\lambda_2$ respectively,
	which can be mathematically described below:
	\begin{align}
	\mu_{i} &= \mu_{i} ^ {l} - (\lambda ^ {l} - \lambda ) \vartheta_{i} ^ {l}, \label{eqn18_2} \\
	\rho_{k}  &= \rho_{k} ^ {l} - (\lambda ^ {l} - \lambda ) \vartheta_{n_\mathrm{A}^l + k} ^ {l}, \label{eqn19_2}
	\end{align}
	where 
	\begin{equation}
	\begin{aligned}
		\bm{\vartheta}^l 
		&= 
		(\mathbf{E}_{\mathrm{AC}}^l)^{-1} 
		\begin{bmatrix}
			\bm{e}_{n_{\mathrm{A}}^l} \\
			(1 - \epsilon) \bm{e}_{n_{\mathrm{C}}^l}
		\end{bmatrix}\\
		&= (\mathbf{E}_{\mathrm{AC}}^l)^{-1}  \mathcolorbox{myhlcolor}{\bm{e}_{n_\mathrm{AC}}^l}.
	\end{aligned}
	\end{equation}
\end{theorem}

\begin{proof}
	\Cref{theorem2} can be proved in two steps, \ie, proving that Lagrangian multipliers $\mu_{i} ~ (i = 1, 2, \cdots, n_{\mathrm{A}}^{l})$ and $\rho_{k} ~ (k = 1, 2, \cdots, n_{\mathrm{C}}^{l})$ are piecewise linear \textit{w.r.t.} the regularization parameter $\lambda_2$ respectively.
	
	First of all, the $l$-th step function of \Cref{eq_f_2} is
	\begin{equation}
		f_2^l(\bm{x}) = 
		\frac{1}{\lambda_2^l} 
		\begin{bmatrix}
			\bm{x}^{\top} &  1
		\end{bmatrix} 
		\mathbf{G}_{\mathrm{inv}}
		(\mathbf{F}^{\top} \bm{\mu}^l
		+
		\mathbf{H}^{\top} \bm{\rho}^l). \label{eq_f_2^l}\\
	\end{equation}

	By the following transformation about $f_2$ and $f_2^l$, we can obtain
	\begin{equation} \label{eq_f_2^l_f_2}
		\begin{aligned}
			f_2(\bm{x}) = &\frac{\lambda_2^l}{\lambda_2}f_2^l(\bm{x}) + f_2(\bm{x})
			- \frac{\lambda_2^l}{\lambda_2}f_2^l(\bm{x}) \\
			= & \frac{1}{\lambda_2} \{ 
			\lambda_2^l f_2^l(\bm{x}) -
			\begin{bmatrix}
				\bm{x}^{\top} & 1
			\end{bmatrix}
			\mathbf{G}_{\mathrm{inv}}
			[\mathbf{F}^{\top} (\bm{\mu}^l - \bm{\mu}) + \mathbf{H}^{\top} (\bm{\rho}^l - \bm{\rho})]
			\}.
		\end{aligned}
	\end{equation}
	For $\forall i \in \mathcal{L}_{\mathrm{A}}^{1, l} \cup \mathcal{R}_{\mathrm{A}}^{1, l}$ and
	$\forall k \in \mathcal{L}_{\mathrm{C}}^{1, l} \cup \mathcal{R}_{\mathrm{C}}^{1, l}$, 
	there are $\mu_i^l - \mu_i = 0$ and $\rho_k^l - \rho_k = 0$ respectively. 
	Therefore, the above equation \eqref{eq_f_2^l_f_2} can be simplified as
	\begin{equation} \label{eq_f_2_system}
		\begin{aligned}
			f_2(\bm{x}) 
			= & \frac{1}{\lambda_2} \{ 
			\lambda_2^l f_2^l(\bm{x}) -
			\begin{bmatrix}
				\bm{x}^{\top} & 1
			\end{bmatrix}
			\mathbf{G}_{\mathrm{inv}}
			[\mathbf{F}_{\mathrm{E}}^{\top} (\bm{\mu}_{\mathrm{E}}^l - \bm{\mu}_{\mathrm{E}}) + \mathbf{H}^{\top} (\bm{\rho}^l_{\mathrm{E}} - \bm{\rho}_{\mathrm{E}})]
			\} \\
			= & \frac{1}{\lambda_2} \{ 
			\lambda_2^l f_2^l(\bm{x}) -
			\begin{bmatrix}
				\bm{x}^{\top} & 1
			\end{bmatrix}
			\mathbf{G}_{\mathrm{inv}}
			\begin{bmatrix}
				(\mathbf{F}_{\mathrm{E}}^l)^{\top} & (\mathbf{H}_{\mathrm{E}}^l)^{\top}
			\end{bmatrix}
			\begin{bmatrix}
				\bm{\mu}_{\mathrm{E}}^l - \bm{\mu}_{\mathrm{E}} \\
				\bm{\rho}_{\mathrm{E}}^l - \bm{\rho}_{\mathrm{E}}
			\end{bmatrix}
			\} 
			.
		\end{aligned}
	\end{equation}
	\hl{
	For $\forall i \in \mathcal{E}_{\mathrm{A}}^{1, l}$ and
	$\forall k \in \mathcal{E}_{\mathrm{C}}^{1, l}$, there  are 
	$f_2(\bm{x}_i) = f_2^l(\bm{x}_i) = 1$ and 
	$f_2(\bm{x}_k) =  f_2^l(\bm{x}_k) = 1 - \epsilon$ respectively. 
	Substitute theses two obtained conditions to \mbox{\Cref{eq_f_2_system}} and combine it with \mbox{\Cref{eq_E_AC}}, the following system of linear equation \textit{w.r.t.} the Lagrangian multipliers $\mathbf{\mu}_{\mathrm{E}}$ and $\mathbf{\rho}_{\mathrm{E}}$ can be deduced.
	}
	\begin{equation}
	\begin{aligned}
		\begin{bmatrix}
			\mathbf{F}_{\mathrm{E}}^l \\ 
			\mathbf{H}_{\mathrm{E}}^l
		\end{bmatrix}
		\mathbf{G}_{\mathrm{inv}}
		\begin{bmatrix}
			(\mathbf{F}_{\mathrm{E}}^l)^{\top} & (\mathbf{H}_{\mathrm{E}}^l)^{\top}
		\end{bmatrix}
		\begin{bmatrix}
			\bm{\mu}_{\mathrm{E}}^l - \bm{\mu}_{\mathrm{E}} \\
			\bm{\rho}_{\mathrm{E}}^l - \bm{\rho}_{\mathrm{E}}
		\end{bmatrix} 
		&= \mathbf{E}_{\mathrm{AC}}^l
		\begin{bmatrix}
			\bm{\mu}_{\mathrm{E}}^l - \bm{\mu}_{\mathrm{E}} \\
			\bm{\rho}_{\mathrm{E}}^l - \bm{\rho}_{\mathrm{E}}
		\end{bmatrix} \\
		&= (\lambda_2^l - \lambda_2) \mathcolorbox{myhlcolor}{\bm{e}_{n_\mathrm{AC}}^l}.
	\end{aligned}
	\end{equation}
	If the matrix $\mathbf{E}_{\mathrm{AC}}^l$ is invertible, then we can obtain
	\begin{equation}
	\begin{aligned}
		\begin{bmatrix}
			\bm{\mu}_{\mathrm{E}}^l - \bm{\mu}_{\mathrm{E}} \\
			\bm{\rho}_{\mathrm{E}}^l - \bm{\rho}_{\mathrm{E}}
		\end{bmatrix} 
		&= (\lambda_2^l - \lambda_2) (\mathbf{E}_{\mathrm{AC}}^l)^{-1} \mathcolorbox{myhlcolor}{\bm{e}_{n_\mathrm{AC}}^l} \\
		&= (\lambda_2^l - \lambda_2) \bm{\vartheta}^l.
	\end{aligned}
	\end{equation}

	In conclusion, \Cref{theorem2} is proved.
\end{proof}

According to \Cref{theorem2}, it can be obtained the following corollary:

\begin{corollary}
	For the QPP (\ref{eqn2_t}), $\lambda_2 ^ {l} > \lambda_2 > \lambda_2 ^ {l + 1}$, the function $f_2$ is piecewise linear \textit{w.r.t.} $\frac{1}{\lambda_2}$.
	\begin{equation}
	\begin{aligned}
		f_2(\bm{x}) 
		&= \frac{1}{\lambda_2}
		[ \lambda_2^l f_2^l(\bm{x}) 
		-
		(\lambda_2^l - \lambda_2)
		h^l(\bm{x})
		] \\
		&= \frac{\lambda_2^l}{\lambda_2} [ f_2^l(\bm{x})- h^l(\bm{x})]
		+ h^l(\bm{x}),
	\end{aligned}
	\end{equation}
	where 
	\begin{equation}
		h^l(\bm{x}) =
		\begin{bmatrix}
			\bm{x}^{\top} & 1
		\end{bmatrix}
		\mathbf{G}_{\mathrm{inv}}
		\begin{bmatrix}
			(\mathbf{F}_{\mathrm{E}}^l)^{\top} & (\mathbf{H}_{\mathrm{E}}^l)^{\top}
		\end{bmatrix}
		\bm{\vartheta}^l.
	\end{equation}
\end{corollary}

\subsection{Initialization} \label{sec_initialization}
When the regularization parameters are infinite, it is easy to prove that the corresponding Lagrangian multipliers are 1. Based on this result, we design an initialization algorithm to get the starting conditions.
\subsubsection{Initializing $\lambda_1^0$} \label{sec_init_1}
The general idea of initialization is to find the value of the Lagrangian multipliers $\bm{\alpha}$ and $\beta$ as the regularization parameter $\lambda_1^0$ approaches infinity, which can be concluded in \Cref{theorem3}.

\begin{theorem} \label{theorem3}
	For the QPP (\ref{eqn3}), when the regularization parameter $\lambda_1^0$ is close to infinity, the Lagrangian multipliers $\alpha_{i}^0 ~ (i = 1, 2, \cdots, n_{\mathrm{B}}^{0})$ and $\beta_{k}^0 ~ (k = 1, 2, \cdots, n_{\mathrm{C}}^{0})$ are equal to 1, \ie,  $\lambda_1^0 \rightarrow + \infty \Rightarrow \alpha_i^0 = \beta_k^0 = 1$,
\end{theorem}

\begin{proof}
	When $\lambda_1^0 \rightarrow +\infty$, we can obtain $- f_1(\bm{x}_i) = 0 < 1 - \epsilon < 1$ from \Cref{eq_f_1}. 
	From \Cref{fign1}, $- f_1(\bm{x}_i) < 1$  is equivalent to 
	$\alpha_i = 1$.
	From \Cref{fign2}, $- f_1(\bm{x}_k) < 1 - \epsilon$  is equivalent to 
	$\beta_k = 1$.
	
	In conclusion, \Cref{theorem3} is proved.
\end{proof}

\subsubsection{Initializing $\lambda_2^0$} \label{sec_init_2}
Accordingly, \Cref{theorem4} is used to find the value of the Lagrangian multipliers $\bm{\mu}$ and $\bm{\rho}$ as the regularization parameter $\lambda_2^0$ approaches the infinity.

\begin{theorem} \label{theorem4}
	For the QPP (\ref{eqn2_t}), when the regularization parameter $\lambda_2^0$ is close to infinity, the Lagrangian multipliers $\mu_{i}^0 ~ (i = 1, 2, \cdots, n_{\mathrm{A}}^{0})$ and $\rho_{k} ~ (k = 1, 2, \cdots, n_{\mathrm{C}}^{0})$ are equal to 1, \ie,  $\lambda_2^0 \rightarrow + \infty \Rightarrow \mu_i^0 = \rho_k^0 = 1$,
\end{theorem}

\begin{proof}
	When $\lambda_2^0 \rightarrow +\infty$, we can obtain 
	$f_1(\bm{x}_i) = 0 < 1 - \epsilon < 1$ according to \Cref{eq_f_1}. 
	From \Cref{fign3}, $f_1(\bm{x}_i) < 1$  is equivalent to 
	$\mu_i = 1$.
	From \Cref{fign4}, $f_1(\bm{x}_k) < 1 - \epsilon$  is equivalent to 
	$\rho_k = 1$.
	
	In conclusion, \Cref{theorem4} is proved.
\end{proof}

\subsubsection{Initialization Algorithm}

According to \Cref{theorem3,theorem4}, it is obvious that Lagrangian multipliers are equal to 1 as the regularization parameter approaches infinity. 
At this point, the samples are all on the left of the elbow.
The basic idea of initialization is to gradually reduce the regularization parameter until the first sample point reaches the elbow. 
\Cref{algorithm1} describes the detailed process of initialization for the QPP \eqref{eqn3}, and it is in the same way for the QPP \eqref{eqn2_t}.

\begin{algorithm*}
	\label{algorithm1}
	\caption{Initialization Algorithm of the QPP \eqref{eqn3}}
	\LinesNumbered
	\KwIn{System parameters $\delta$, $\epsilon$ and sample matrices $\mathbf{A}$, $\mathbf{B}$ and $\mathbf{C}$.}
	\KwOut{Initial parameters $\lambda_1 ^ {0}$, initial Lagrangian multipliers $\alpha_{i} ^ {0}$, $\beta_{k} ^ {0}$, and initial index sets $\mathcal{L}_{\mathrm{B}} ^ {1, 0}$, $\mathcal{E}_{\mathrm{B}} ^ {1, 0}$, $\mathcal{R}_{\mathrm{B}} ^ {1, 0}$, $\mathcal{L}_{\mathrm{C}} ^ {1, 0}$, $\mathcal{E}_{\mathrm{C}} ^ {1,0}$ and $\mathcal{R}_{\mathrm{C}} ^ {1, 0}$.}
	\tcp{Variable preparation phase}
	$\bm{\alpha}^0 \leftarrow \bm{e}_{n_{\mathrm{B}}}$, 
	$n_{\mathrm{B}} \leftarrow \mathrm{size}(\mathbf{B}, 1)$\tcp*{By \Cref{theorem3}.}
	$\bm{\beta}^0 \leftarrow \bm{e}_{n_{\mathrm{C}}}$, 
	$n_{\mathrm{B}} \leftarrow \mathrm{size}(\mathbf{C}, 1)$\tcp*{By \Cref{theorem3}.}
	Obtain homogeneous matrices $\mathbf{F}$, $\mathbf{G}$ and $\mathbf{H}$ from $\mathbf{A}$, $\mathbf{B}$ and $\mathbf{C}$\;
	$\mathbf{I} \leftarrow \mathrm{eye}(m + 1)$, 
	$m \leftarrow \mathrm{size}(\mathbf{A}, 1)$\;
	$\mathbf{F}_{\mathrm{inv}} \leftarrow (\mathbf{F}^{\top} \mathbf{F} + \delta \mathbf{I})^{-1}$
	\tcp*{By \Cref{eq_F_inv}}
	$\lambda_1^0 \leftarrow 0$\tcp*{Initialize a minimum number to $\lambda_1^0$}
	$\mathcal{R}_{\mathrm{B}} ^ {1, 0}$, $\mathcal{R}_{\mathrm{C}} ^ {1, 0} \gets \varnothing,\varnothing$\;
	\tcp{Initialization phase for samples in $\mathcal{B}$: see \Cref{sec_init_1}}
	\ForEach{$i \in \mathcal{B}$} {
	$
	\lambda =  
	\begin{bmatrix}
		\bm{x}_i^{\top} &  1
	\end{bmatrix} 
	\mathbf{F}_{\mathrm{inv}}
	(\mathbf{G}^{\top} \bm{\alpha}^0
	+
	\mathbf{H}^{\top} \bm{\beta}^0)$\tcp*{By \Cref{eq_f_1}}
		\If{$\lambda > \lambda_1^0$}{
			$\lambda_1^0 \leftarrow \lambda$\tcp*{Update the regularization parameter by}
			\hl{$\mathcal{L}_{\mathrm{B}} ^ {1, 0}, \mathcal{E}_{\mathrm{B}} ^ {1, 0} \gets \{{1, \cdots, i-1, i+1, \cdots, n_{\mathrm{B}}}\}, \{i\}$} \tcp*{Initialize samples in $\mathcal{B}$ when $\mathcal{L}_{\mathrm{B}} \rightarrow \mathcal{E}_{\mathrm{B}}$}
			\hl{$\mathcal{L}_{\mathrm{C}} ^ {1, 0}, \mathcal{E}_{\mathrm{C}} ^ {1, 0} \gets \{{1, 2, \cdots, n_{\mathrm{B}}}\}, \varnothing$} \tcp*{Initialize samples in $\mathcal{C}$}
		}
	}
	\tcp{Initialization phase for samples in $\mathcal{C}$: see \Cref{sec_init_2}}
	\ForEach{$k \in \mathcal{C}$} {
	$
	\lambda = 
	\frac{1}{1 - \epsilon}
	\begin{bmatrix}
		\bm{x}_k^{\top} &  1
	\end{bmatrix} 
	\mathbf{F}_{\mathrm{inv}}
	(\mathbf{G}^{\top} \bm{\alpha}^0
	+
	\mathbf{H}^{\top} \bm{\beta}^0)$\tcp*{By \Cref{eq_f_1}}
		\If{$\lambda > \lambda_1^0$}{
			$\lambda_1^0 \leftarrow \lambda$\tcp*{Update the regularization parameter by}
			\hl{$\mathcal{L}_{\mathrm{C}} ^ {1, 0}, \mathcal{E}_{\mathrm{C}} ^ {1, 0} \gets \{{1, \cdots, i-1, i+1, \cdots, n_{\mathrm{C}}}\}, \{i\}$} \tcp*{Initialize samples in $\mathcal{C}$ when $\mathcal{L}_{\mathrm{C}} \rightarrow \mathcal{E}_{\mathrm{C}}$}
			\hl{$\mathcal{L}_{\mathrm{B}} ^ {1, 0}, \mathcal{E}_{\mathrm{B}} ^ {1, 0} \gets \{{1, 2, \cdots, n_{\mathrm{B}}}\}, \varnothing$} \tcp*{Initialize samples in $\mathcal{B}$}
		}
	}
	\Return $\lambda_1 ^ {0}$, $\alpha_{i} ^ {0}$, $\beta_{k} ^ {0}$, $\mathcal{L}_{\mathrm{B}} ^ {1, 0}$, $\mathcal{E}_{\mathrm{B}} ^ {1, 0}$, $\mathcal{R}_{\mathrm{B}} ^ {1, 0}$, $\mathcal{L}_{\mathrm{C}} ^ {1, 0}$, $\mathcal{E}_{\mathrm{C}} ^ {1,0}$ and $\mathcal{R}_{\mathrm{C}} ^ {1, 0}$\;
\end{algorithm*}

\subsection{Finding $\lambda ^ {l + 1}$}
The aim of finding $\lambda ^ {l + 1}$ is to update the regularization parameter as long as the corresponding parameters, such as sample sets and Lagrangian multipliers. Before designing the updating strategy, we first give the definition of the event and starting event, as follows:

\begin{definition}{} \label{def_1}
	As the regularization parameter changes, the sample index set will also be updated, and this change is defined as an \textbf{event} in this work, represented by a symbolic "$\rightarrow$".
\end{definition}

\begin{definition}{}
	If more than one event can occur, the event with the largest regularization parameter value is selected to occur first and is defined as the \textbf{starting event} in this work.
\end{definition}


\begin{figure}
	\centering
	\includegraphics[width=\linewidth]{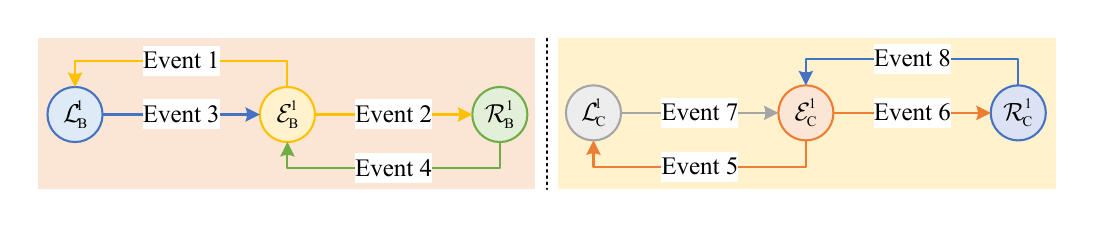}
	\caption{Possible events for the QPP \eqref{eqn3}: there are two scenarios and eight events for the sample sets $\mathcal{B}$ (left) and $\mathcal{C}$ (right).}
	\label{fig_possible_events}
\end{figure}

As analyzed in \Cref{sec_partition_strategy}, the changes of samples in sets $\mathcal{B}$ and $\mathcal{C}$ need to be discussed for the QPP \eqref{eqn3}. From \Cref{def_1}, there are eight kinds of possible events in total, illustrated in \Cref{fig_possible_events}. Next, we elaborate on them one by one.

If $\mathcal{E}_{\mathrm{B}} ^ {1} \neq \varnothing$,
then one of the sample points in $\mathcal{E}_{\mathrm{B}} ^ {1}$ may go into either $\mathcal{L}_{\mathrm{B}} ^ {1}$ or $\mathcal{R}_{\mathrm{B}} ^ {1}$.

\begin{enumerate}[\bf{Event 1}]
	\item If $\mathcal{E}_{\mathrm{B}} ^ {1} \rightarrow \mathcal{L}_{\mathrm{B}} ^ {1}$, 
	then $\alpha_i$ changes from $0 \leq \alpha_{i} \leq 1$ to $\alpha_{i} = 1$ and $f_{1} (\bm{x}_{i})$ changes from $-f_{1} (\bm{x}_{i}) = 1$ to  $-f_{1} (\bm{x}_{i}) < 1$. \hl{Combining $\alpha_i = 1$ with \mbox{\Cref{eqn18}}, we can obtain the corresponding regularization parameter $\lambda_1^1$ by} 
\begin{equation}
\label{eq_event_1_1}
\lambda_{1}^{1} 
= \max_{i \in \mathcal{E} ^ {1, l}_{\mathrm{B}}} \{ \lambda_1 ^ {l} - \frac{\alpha ^ {l}_{i} - \alpha_{i}} {\theta ^ {l}_{i}}\}
= \max_{i \in \mathcal{E} ^ {1, l}_{\mathrm{B}}} \{ \lambda_1 ^ {l} - \frac{\alpha ^ {l}_{i} - 1}{\theta ^ {l}_{i}}\},
\end{equation}
where $\theta ^ {l}_{i} < 0$.
\end{enumerate}

\begin{enumerate}[\bf{Event 2}]
	\item If $\mathcal{E}_{\mathrm{B}} ^ {1} \rightarrow \mathcal{R}_{\mathrm{B}} ^ {1}$, 
	then $\alpha_i$ changes from $0 \leq \alpha_{i} \leq 1$ to $\alpha_{i} = 0$ and $f_{1} (\bm{x}_{i})$ changes from $-f_{1} (\bm{x}_{i}) = 1$ to  $-f_{1} (\bm{x}_{i}) > 1$. \hl{Combining $\alpha_i = 0$ with \mbox{\Cref{eqn18}}, we can obtain the corresponding regularization parameter $\lambda_1^2$ by}
\begin{equation}
\label{eq_event_1_2}
\lambda_{1}^2
= \max_{i \in \mathcal{E} ^ {1, l}_{\mathrm{B}}} \{\lambda_1 ^ {l} - \frac{\alpha ^ {l}_{i} - \alpha_{i}}{\mathcolorbox{myhlcolor}{\theta ^ {l}_{i}}}\}
= \max_{i \in \mathcal{E} ^ {1, l}_{\mathrm{B}}} \{\lambda_1 ^ {l} - \frac{\alpha ^ {l}_{i}}{\theta ^ {l}_{i}}\},
\end{equation}
where $\theta ^ {l}_{i} > 0$.
\end{enumerate}	  

If $\mathcal{L}_{\mathrm{B}} ^ {1} \neq \varnothing$, then one of the sample points in $\mathcal{L}_{\mathrm{B}} ^ {1}$ may go into $\mathcal{E}_{\mathrm{B}} ^ {1}$.

\begin{enumerate}[\bf{Event 3}]
	\item If $\mathcal{L}_{\mathrm{B}} ^ {1} \rightarrow \mathcal{E}_{\mathrm{B}} ^ {1}$, 
	then $\alpha_i$ changes from $\alpha_{i} = 1$ to $0 \leq \alpha_{i} \leq 1$ and $f_{1} (\bm{x}_{i})$ changes from $-f_{1} (\bm{x}_{i}) < 1$ to  $-f_{1} (\bm{x}_{i}) = 1$. \hl{Combining $-f_{1} (\bm{x}_{i}) = 1$ with \mbox{\Cref{eq_fx_gx}}, we can obtain the corresponding regularization parameter $\lambda_1^3$ by}
\begin{equation}
\label{eq_event_1_3}
\begin{aligned}
\lambda_{1}^3
&= \max_{i \in \mathcal{L} ^ {1, l}_{\mathrm{B}}} \{ \lambda_1 ^ {l} \frac{f_{1} ^ {l} (\bm{x}_{i}) + g ^ {l}(\bm{x}_{i})} {f_{1} (\bm{x}_{i}) + g ^ {l} (\bm{x}_{i})}\} \\
&= \max_{i \in \mathcal{L} ^ {1, l}_{\mathrm{B}}} \{ \lambda_1 ^ {l} \frac{f_{1} ^ {l} (\bm{x}_{i}) + g ^ {l}(\bm{x}_{i})} {-1 + g ^ {l} (\bm{x}_{i})}\}.			
\end{aligned}
\end{equation}
\end{enumerate}

If $\mathcal{R}_{\mathrm{B}} ^ {1} \neq \varnothing$,
then one of the sample points in $\mathcal{R}_{\mathrm{B}} ^ {1}$ may go into $\mathcal{E}_{\mathrm{B}} ^ {1}$.
\begin{enumerate}[\textbf{Event 4}]
	\item If $\mathcal{R}_{\mathrm{B}} ^ {1} \rightarrow \mathcal{E}_{\mathrm{B}} ^ {1}$, 
	then $\alpha$ changes from $\alpha_{i} = 0$ to $0 \leq \alpha_{i} \leq 1$ and $f_{1} (\bm{x}_{i})$ changes from $-f_{1} (\bm{x}_{i}) > 1$ to  $-f_{1} (\bm{x}_{i}) = 1$. \hl{Combining $-f_{1} (\bm{x}_{i}) = 1$ with \mbox{\Cref{eq_fx_gx}}, we can obtain the corresponding regularization parameter $\lambda_1^4$ by}
\begin{equation}
\label{eq_event_1_4}
\begin{aligned}
\mathcolorbox{myhlcolor}{\lambda_1^{4}}
&= \max_{i \in \mathcal{R} ^ {1, l}_{\mathrm{B}}} \{ \lambda_1 ^ {l} \frac{f_{1} ^ {l} (\bm{x}_{i}) + g ^ {l}(\bm{x}_{i})} {f_{1} (\bm{x}_{i}) + g ^ {l} (\bm{x}_{i})} \} \\
&= \max_{i \in \mathcal{R} ^ {1, l}_{\mathrm{B}}} \{ \lambda_1 ^ {l} \frac{f_{1} ^ {l} (\bm{x}_{i}) + g ^ {l}(\bm{x}_{i})} {-1 + g ^ {l} (\bm{x}_{i})} \}.			
\end{aligned}
\end{equation}

\end{enumerate}

If $\mathcal{E}_{\mathrm{C}} ^ {1} \neq \varnothing$,
then one of the sample points in $\mathcal{E}_{\mathrm{C}} ^ {1}$ may go into either $\mathcal{L}_{\mathrm{C}} ^ {1}$ or $\mathcal{R}_{\mathrm{C}} ^ {1}$.
\begin{enumerate}[\textbf{Event 5}]
	\item If $\mathcal{E}_{\mathrm{C}} ^ {1} \rightarrow \mathcal{L}_{\mathrm{C}} ^ {1}$, 
	then $\beta$ changes from $0 \leq \beta_{i} \leq 1$ to $\beta_{i} = 1$ and $f_{1} (\bm{x}_{k})$ changes from $-f_{1} (\bm{x}_{k}) = 1 - \epsilon$ to  $-f_{1} (\bm{x}_{k}) < 1 - \epsilon$. \hl{Combining $\beta_{i} = 1$ with \mbox{\Cref{eqn19}}, we can obtain the corresponding regularization parameter $\lambda_1^5$ by}
\begin{equation}
\label{eq_event_1_5}
\begin{aligned}
\lambda_{1}^5 
&= \max_{k \in \mathcal{E} ^ {1, l}_{\mathrm{C}}} \{ \lambda_1 ^ {l} - \frac{\beta ^ {l}_{k} - \beta_{k}}{\theta ^ {l}_{n_{\mathrm{B}}^l + k}}\}\\
&= \max_{k \in \mathcal{E} ^ {1, l}_{\mathrm{C}}} \{ \lambda_1 ^ {l} - \frac{\beta ^ {l}_{k} - 1}{\theta ^ {l}_{n_{\mathrm{B}}^l + k}}\},
\end{aligned}	
\end{equation}
where $\theta ^ {l}_{n_{\mathrm{B}}^l + k} > 0$.
\end{enumerate}

\begin{enumerate}[\textbf{Event 6}]
	\item If $\mathcal{E}_{\mathrm{C}} ^ {1} \rightarrow \mathcal{R}_{\mathrm{C}} ^ {1}$, 
	then $\beta$ changes from $0 \leq \beta_{i} \leq 1$ to $\beta_{i} = 0$ and $f_{1} (\bm{x}_{k})$ changes from $-f_{1} (\bm{x}_{k}) = 1 - \epsilon$ to  $-f_{1} (\bm{x}_{k}) > 1 - \epsilon$. \hl{Combining $\beta_{i} = 0$ with \mbox{\Cref{eqn19}}, we can obtain the corresponding regularization parameter $\lambda_1^6$ by}
\begin{equation}
\label{eq_event_1_6}
\begin{aligned}
\lambda_{1}^6 
&= \max_{k \in \mathcal{E} ^ {1, l}_{\mathrm{C}}} \{ \lambda_1 ^ {l} - \frac{\beta ^ {l}_{k} - \beta_{k}} {\theta ^ {l}_{n_{\mathrm{B}}^l + k}} \}\\
&= \max_{k \in \mathcal{E} ^ {1, l}_{\mathrm{C}}} \{ \lambda_1 ^ {l} - \frac{\beta ^ {l}_{k}} {\theta ^ {l}_{n_{\mathrm{B}}^l + k}} \},
\end{aligned}
\end{equation}
where $\theta ^ {l}_{n_{\mathrm{B}}^l + k} < 0$.
\end{enumerate}

If $\mathcal{L}_{\mathrm{C}} ^ {1} \neq \varnothing$,
then one of the sample points in $\mathcal{L}_{\mathrm{C}} ^ {1}$ may go into $\mathcal{E}_{\mathrm{C}} ^ {1}$.
\begin{enumerate}[\textbf{Event 7}]
	\item If $\mathcal{L}_{\mathrm{C}} ^ {1} \rightarrow \mathcal{E}_{\mathrm{C}} ^ {1}$, 
	then $\beta$ changes from $\beta_{i} = 1$ to $0 \leq \beta_{i} \leq 1$ and $f_{1} (\bm{x}_{k})$ changes from $-f_{1} (\bm{x}_{k}) < 1 - \epsilon$ to  $-f_{1} (\bm{x}_{k}) = 1 - \epsilon$. \hl{Combining $-f_{1} (\bm{x}_{k}) = 1 - \epsilon$ with \mbox{\Cref{eq_fx_gx}}, we can obtain the corresponding regularization parameter $\lambda_1^7$ by}
\begin{equation}
\label{eq_event_1_7}
\begin{aligned}
\lambda_{1}^7 
&= \max_{k \in \mathcal{L} ^ {1, l}_{\mathrm{C}}} \{\lambda_1 ^ {l} \frac{f_{1} ^ {l}(\bm{x}_{k}) + g ^ {l}(\bm{x}_{k})} {f_{1} (\bm{x}_{k}) + g ^ {l} (\bm{x}_{k})} \} \\
&= \max_{k \in \mathcal{L} ^ {1, l}_{\mathrm{C}}} \{\lambda_1 ^ {l} \frac{f_{1} ^ {l}(\bm{x}_{k}) + g ^ {l}(\bm{x}_{k})} {-(1 - \epsilon) + g ^ {l} (\bm{x}_{k})} \}.
\end{aligned}		
\end{equation}
\end{enumerate}

If $\mathcal{R}_{\mathrm{C}} ^ {1} \neq \varnothing$,
then one of the sample points in $\mathcal{R}_{\mathrm{C}} ^ {1}$ may go into $\mathcal{E}_{\mathrm{C}} ^ {1}$.
\begin{enumerate}[\textbf{Event 8}]
	\item If $\mathcal{R}_{\mathrm{C}} ^ {1} \rightarrow \mathcal{E}_{\mathrm{C}} ^ {1}$, 
	then $\beta$ changes from $\beta_{i} = 0$ to $0 \leq \beta_{i} \leq 1$ and $f_{1} (\bm{x}_{k})$ changes from $-f_{1} (\bm{x}_{k}) > 1 - \epsilon$ to  $-f_{1} (\bm{x}_{k}) = 1 - \epsilon$. \hl{Combining $-f_{1} (\bm{x}_{k}) = 1 - \epsilon$ with \mbox{\Cref{eq_fx_gx}}, we can obtain the corresponding regularization parameter $\lambda_1^8$ by}
\begin{equation}
\label{eq_event_1_8}
\begin{aligned}
\mathcolorbox{myhlcolor}{\lambda_1^{8}}
&= \max_{k \in \mathcal{R} ^ {1, l}_{\mathrm{C}}} \{\lambda_1 ^ {l} \frac{f_{1} ^ {l}(\bm{x}_{k}) + g ^ {l}(\bm{x}_{k})} {f_{1} (\bm{x}_{k}) + g ^ {l} (\bm{x}_{k})} \} \\
&= \max_{k \in \mathcal{R} ^ {1, l}_{\mathrm{C}}} \{\lambda_1 ^ {l} \frac{f_{1} ^ {l}(\bm{x}_{k}) + g ^ {l}(\bm{x}_{k})} {- (1 - \epsilon) + g ^ {l} (\bm{x}_{k})} \}.
\end{aligned}
\end{equation}
\end{enumerate}

\hl{When the regularization parameter decreases as the iteration step size increases, eight possible events in the above six cases are discussed above. One of these events that occurs is called as the starting event. The criteria for selecting the starting event is that when this event occurs, the corresponding regularization parameter is at the maximum among them and simultaneously greater than the minimum threshold $\lambda_{\mathrm{min}}$. Thus, the regularization parameter $\lambda ^ {l + 1}$ and the corresponding starting event $e_{\mathrm{start}}$ can be calculated the following equations, respectively.}
\begin{eqnarray}
\lambda_1  &=& \max \{ \mathcolorbox{myhlcolortwo}{\lambda_{1}^1}, \lambda_{1}^2, \cdots, \lambda_{1}^8 \}, \label{eq_lambda_l+1}\\
e_{\mathrm{start}} &=& \mathop{\arg \max}_{i = 1, 2, \cdots, 8}\{ \mathcolorbox{myhlcolortwo}{\lambda_{1}^1}, \lambda_{1}^2, \cdots, \lambda_{1}^8 \}. \label{eq_argmax}
\end{eqnarray}

In this way, the $(l+1)$ step parameters are updated according to the starting event $e_{\mathrm{start}}$. Then the next iteration proceeds until the regularization parameter $\lambda_1$ approaches the minimum threshold, which is elaborated in \Cref{algorithm2}.

\begin{figure*}
\tiny


	\includegraphics[width=\linewidth]{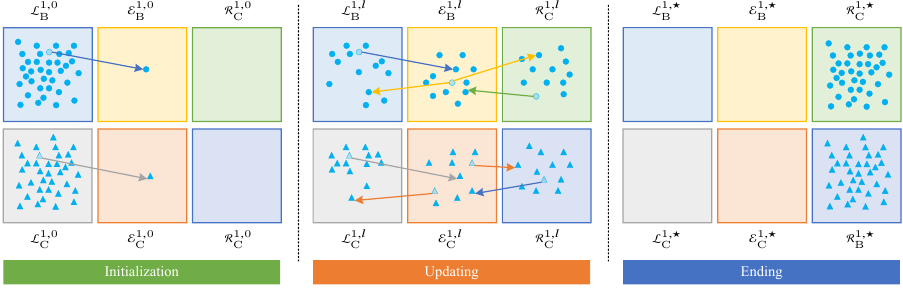}
	\caption{Illustration of our proposed solution path algorithm for the QPP \eqref{eqn2_t}: According to \Cref{theorem3}, all of samples in $\mathcal{B}$ initially locate in $\mathcal{L}_{\mathrm{B}}$ and $\mathcal{L}_{\mathrm{C}}$ when the regularization parameter $\lambda_1$ approaches infinity. Left: invoke \Cref{algorithm1} to initialize the algorithm and obtain the initial index sets $\mathcal{L}_{\mathrm{B}}^{1,0}, \mathcal{E}_{\mathrm{B}}^{1,0}, \mathcal{R}_{\mathrm{B}}^{1,0}, \mathcal{L}_{\mathrm{C}}^{1,0}, \mathcal{E}_{\mathrm{C}}^{1,0}, \mathcal{R}_{\mathrm{C}}^{1,0}$. Middle: invoke \Cref{algorithm2} to update these parameters until the termination condition ends. Right: The ideal ending is that all of samples are located in $\mathcal{R}_{\mathrm{B}}$ and $\mathcal{R}_{\mathrm{C}}$.}
	\label{fig_framework}
\end{figure*}

For the QPP \eqref{eqn2_t}, it is also similar to the above.

\subsection{Solution Path Algorithm} \label{sec_algorithm}
Our proposed solution path algorithm is illustrated in \Cref{fig_framework}.
For the sample set with $K (K > 3)$ classes, to obtain the entire solution path of regularization parameters $\lambda$, we need the following three steps.

\textbf{Step 1}: The data set is randomly divided into training set and test set, and the sampling rate of test samples is $r$. 

\textbf{Step 2}: Any two categories of samples $K_i$ and $K_j ~ (i \neq j) $ are randomly selected from the training sample set, denoted as class $+1$ and $-1$, respectively, and the rest are denoted as class $0$. Sets of the three classes of samples are represented by $\mathcal{A}$, $\mathcal{B}$ and $\mathcal{C}$. Therefore, there are $K(K - 2) / 2$ combinations for samples with $K$ categories, and the optimal hyperplanes $f_{1, i, j}(\bm{x})$ and $f_{2, i, j}(\bm{x})$ corresponding to each combination $(K_i, K_j)$ are calculated respectively.

\textbf{Step 2.1}: For the QPP \eqref{eqn3}, there are two steps to the solution.
\begin{itemize}
	\item Initialization: Set the initial value of system parameters $\delta$ and $\epsilon$, then invoke \Cref{algorithm1} to determine the initial regularization parameter $\lambda_1 ^ {0}$, the initial Lagrangian multipliers $\alpha_{i} ^ {0}$, $\beta_{k} ^ {0}$, the initial index sets $\mathcal{L}_{\mathrm{B}} ^ {1, 0}$, $\mathcal{E}_{\mathrm{B}} ^ {1, 0}$, $\mathcal{R}_{\mathrm{B}} ^ {1, 0}$, $\mathcal{L}_{\mathrm{C}} ^ {1, 0}$, $\mathcal{E}_{\mathrm{C}} ^ {1,0}$ and $\mathcal{R}_{\mathrm{C}} ^ {1, 0}$.

	\item Updating: Invoke \Cref{algorithm2} to get the entire solution to the QPP \eqref{eqn3}.
\end{itemize}

\textbf{Step 2.2}: For the QPP \eqref{eqn2_t}, there are two steps to the solution similar to the QPP \eqref{eqn3}.

\begin{algorithm*}
	\label{algorithm2}
	\caption{Traversal Search Algorithm of the QPP \eqref{eqn3}}
	\LinesNumbered
	\KwIn{System parameters $\lambda_{\min}$ and $l_{\max}$, initial regularization parameter $\lambda_1 ^ {0}$, initial Lagrangian multipliers $\alpha_{i} ^ {0}$, $\beta_{k} ^ {0}$, initial index sets $\mathcal{L}_{\mathrm{B}} ^ {1, 0}$, $\mathcal{E}_{\mathrm{B}} ^ {1, 0}$, $\mathcal{R}_{\mathrm{B}} ^ {1, 0}$, $\mathcal{L}_{\mathrm{C}} ^ {1, 0}$, $\mathcal{E}_{\mathrm{C}} ^ {1,0}$ and $\mathcal{R}_{\mathrm{C}} ^ {1, 0}$.}
	\KwOut{Solution path $\bm{\lambda}_1$, $\bm{\alpha}$ and $\bm{\beta}$.}
	$l \gets 0$\;
	\While {$\lambda_1 ^ {l} \geq \lambda_{\min} ~ \mathbf{and} ~ l \leq l_{\max}$  } {
		$n_{\mathrm{B}}^l \gets \mathrm{length} (\mathcal{E}_{\mathrm{B}} ^ {1, l})$,
		$n_{\mathrm{C}}^l \gets \mathrm{length} (\mathcal{E}_{\mathrm{C}} ^ {1, l})$\tcp*{Number of  samples in the elbow}
		Calculate $\bm{\theta}^l$ according to \Cref{eq_theta_l}\;
		\tcp{Traverses through 8 defined events to see if they occur}
		\ForEach{$e \in \{1,2,\cdots,8\}$}{
			\eIf{event $e$ occurs}{
				\hl{Obtain $\lambda_1^e$ according to \mbox{\Cref{eq_event_1_1,eq_event_1_2,eq_event_1_3,eq_event_1_4,eq_event_1_5,eq_event_1_6,eq_event_1_7,eq_event_1_8}}\;}
			}{
				$\lambda_1^e \gets 0$\tcp*{Assign 0 if event $e$ doesn't occur}
			}
		}
		\tcp{Update the iteration and the $(l+1)$-step parameters}
		$\lambda_1^{l+1} \gets \max\{\lambda_1^1, \lambda_1^2, \cdots, \lambda_1^8\}$\tcp*{Obtain the regularization parameter by \Cref{eq_lambda_l+1}}
			$e_{\mathrm{start}} \gets \mathop{\arg \max}\{\lambda_1^1, \lambda_1^2, \cdots, \lambda_1^8\}$\tcp*{Determine which event occurs by \Cref{eq_argmax}}
			Update $\mathcal{L}_{\mathrm{B}} ^ {1, l + 1},\mathcal{E}_{\mathrm{B}} ^ {1, l + 1},\mathcal{R}_{\mathrm{B}} ^ {1, l + 1},\mathcal{L}_{\mathrm{C}} ^ {1, l + 1},\mathcal{E}_{\mathrm{C}} ^ {1, l + 1},\mathcal{R}_{\mathrm{C}} ^ {1, l + 1}$ according to the event $e_{\mathrm{start}}$;\\
			Update the Lagrangian multipliers $\alpha_{i} ^ {l + 1}$, $\beta_{k} ^ {l + 1}$ according to \Cref{theorem1}\;
			$l \gets l + 1$\;
	}
	\Return $\bm{\lambda}_1$, $\bm{\alpha}$ and $\bm{\beta}$;
\end{algorithm*}

\textbf{Step 3}: To determine the optimal regularization parameter pairs $(\lambda, ~\bar{\lambda})$ and its corresponding hyperplanes $f_{1, i, j}(\bm{x})$ and $f_{2, i, j}(\bm{x})$, the decision function $f_{i,j}(\bm{x})$ is utilized to evaluate the error for every combination $(K_i, K_j)$ on the examining set, and it generates ternary outputs $\{1, 0, -1\}$.
\begin{equation}
\label{eqn61}
f_{i,j}(\bm{x}) = \left \{
\begin{aligned}
1, ~& \mathrm{if} ~ f_{1, i, j}(\bm{x}) > -1 + \epsilon, \\
-1, ~& \mathrm{if} ~ f_{2, i, j}(\bm{x}) < 1 - \epsilon , \\
0, ~& \mathrm{otherwise}.
\end{aligned}
\right.
\end{equation}

\textbf{Step 4}: Predict the classes of samples in the test set using $K(K - 2) / 2$ pairs of the optimal regularization parameters. For any sample $\bm{x}$ in test set, we need to calculate its votes under the decision function $f_{i,j}(\bm{x})$. 
\begin{itemize}
	\item If $f_{1, i, j}(\bm{x}) > -1 + \epsilon$, then class $K_{i}$ gets ``1'' vote. 
	\item If $f_{2, i, j}(\bm{x}) < 1 - \epsilon$ and $f_{1, i, j}(\bm{x}) < -1 + \epsilon$, then class $K_{j}$ gets ``1'' vote.
	\item If $f_{1, i, j}(\bm{x}) \leq -1 + \epsilon$ or $f_{2, i, j}(\bm{x}) \geq 1 - \epsilon$, then both classes get ``0'' vote. 
	\item If $f_{1, i, j}(\bm{x}) > -1 + \epsilon$ and $f_{2, i, j}(\bm{x}) < 1 - \epsilon$, then both classes get ``$-1$'' vote.
\end{itemize}
Calculate the votes of $K$ classes of the sample under $K(K - 2) / 2$ decision functions, and the class with the most votes is its final category.

\begin{table*}
\small
	\renewcommand{\arraystretch}{1.3}
	\caption{Description of data sets used in this work.}
	\label{Tab_1}
	\centering
	
	\begin{tabular}{cccccc}
		\toprule
		\bfseries No. & 
		\bfseries Data sets & 
		\bfseries \# Tol.$^ {\rm{a}}$ &
		\bfseries \# D$^ {\rm{b}}$ & 
		\bfseries \# K$^ {\rm{c}}$ &
		\bfseries \# Size of classes\\
		\midrule
 1 & Balancescale & 625 & 4 & 3 & 49, 288, 288 \\
 2 & CMC & 1473 & 9 & 3 & 629, 333, 511 \\
 3 & Glass & 214 & 9 & 6 & 29, 76, 70, 17, 13, 9 \\
 4 & Iris & 150 & 4 & 3 & 50, 50, 50 \\
 5 & Robotnavigation & 5456 & 24 & 4 & 826, 2097, 2205, 328 \\
 6 & Seeds & 210 & 7 & 3 & 70, 70, 70 \\
 7 & Thyroid & 215 & 5 & 3 & 150, 35, 30 \\
 8 & Vowel & 871 & 3 & 6 & 72, 89, 172, 151, 207, 180 \\
 9 & Wine & 178 & 13 & 3 & 59, 71, 48\\
		\toprule
	\end{tabular}
	\begin{threeparttable}
		\begin{tablenotes}
			\item[a] The total number of instances in the data set.
			\item[b] The dimension of the features of the instance in the data set.
			\item[c] The number of instance classes in the data set.
		\end{tablenotes}
	\end{threeparttable}
\end{table*}

\section{Numerical Experiments} \label{Experiments}

In this section, we first verify the piecewise linear theory in \Cref{theorem1} and \ref{theorem2} with experiments. We further test the proposed algorithm on prediction accuracy and training time in nine different data sets. The computational overhead and time complexity of the proposed algorithm are finally discussed.

\subsection{Data Sets}
According to the data sets in the relevant works \citep{Xu2013A, DingReview}, we have selected nine data sets with different feature dimensions, number of categories and total number of instances to evaluate the performance of the proposed algorithm.
All the data sets can be achieved from the UCI machine learning repository\footnote{UCI machine learning repo]: https://archive.ics.uci.edu/ml/index.php}, including Balancescale, CMC, Dermatology, Glass, Iris, Seeds, Thyroid, Vowel and Wine. 
And the detailed description, such as the size of classes of these data sets, is summarized in \Cref{Tab_1}. 
The main goal of our work is to save parameter tuning time, so we deleted a few classes for some data sets. For instance, the original number of each class of data set Ecoli is 143, 77, 2, 2, 35, 20, 5 and 52 respectively. To reduce the cross-validation time, we can delete the classes with 3, 4 and 8 instances in the experiments.

\subsection{Implementation Details}

We compare against TSVM baselines with different strategies as in OVOVR TSVM \citep{Xu2013A}, OVR TSVM \citep{cong2008efficient} and \hlfour{OVO TSVM \mbox{\citep{DingReview}}}.  We further compare against SVM baselines  with different strategies as in OVR SVM \citep{AnguloK} and OVO SVM \citep{shieh2008multiclass}. \hlfour{Note that all baselines are implemented based on the idea of multi-class strategies and the original TSVM/SVM.}
We closely follow the experimental setting in \citep{Xu2013A}. 
On the large data sets, 10-fold cross-validation method is used to find the optimal parameters. Otherwise, we adopt leave-one method for validation. 
In the experiment, we set $r = 0.25$, $\delta = 10^{-4}$, $\epsilon = 0.05$, $l_{\max} = 1000$ and $\lambda_{\min} = 10^{-4}$.
For the vote strategy, we use two different decision functions to test ours and OVOVR TSVM.
Furthermore, all the experiments are repeated ten times using the same parameter configuration. 

For the grid search method, suppose the regularization parameter deduces from $\lambda = 1000$, and the step is set to $\Delta \lambda = 0.1$. We use the `quadprog.m' function to realize the QPP in MATLAB. 

All the experiments  are performed by MATLAB R2016b on a personal computer equipped with an Intel (R) Core (TM) i7-7500U 2.90 GHz CPU and 8 GB memory capacity.

\subsection{Results and Analysis}
We first visualize a classification process and an entirely regularized solution path of two sub-optimization problems to verify the classification and pairwise linear theory, respectively. Then, we analyze the first event and compare the prediction accuracy performance and training time with state-of-the-art methods. Finally, we discuss the computational overhead and time complexity of TSVMPath.

\subsubsection{Case Study}

\begin{figure*}
\centering
	\subfigure[Original Data]{\includegraphics[width=0.32\linewidth]{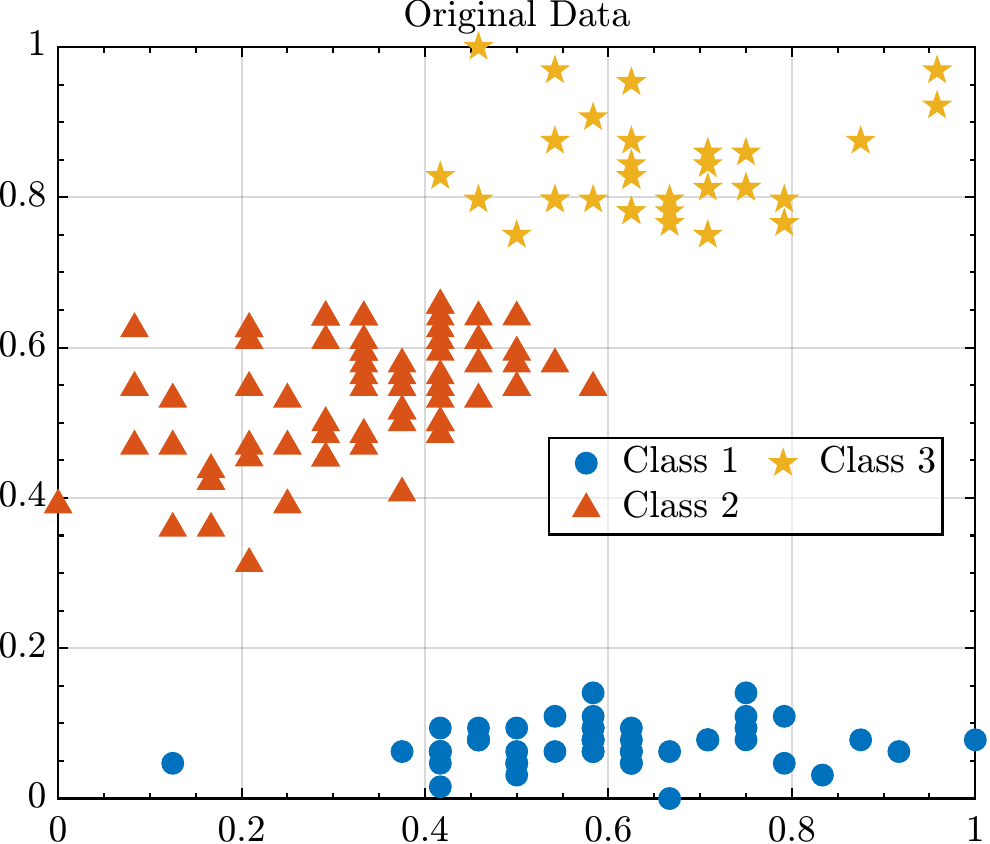} \label{fig_plot_example_orig}}
	\subfigure[Prediction (Ours)]{\includegraphics[width=0.32\linewidth]{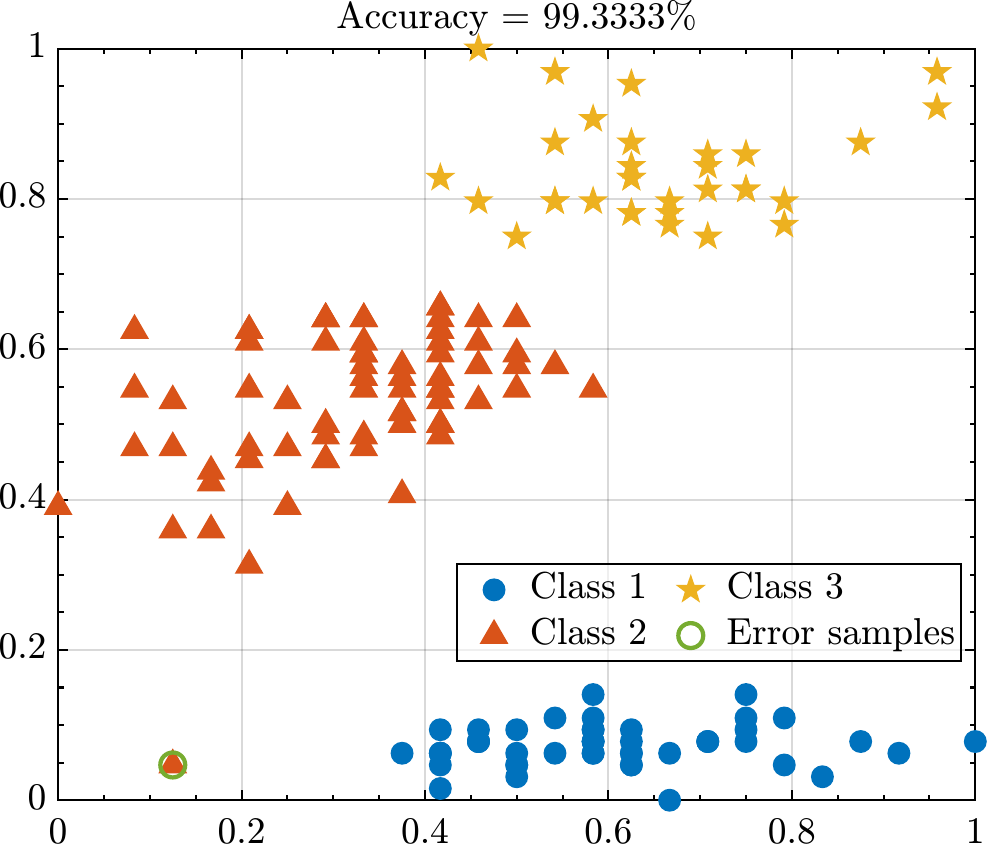} \label{fig_plot_example_pred}}
	\subfigure[Prediction (OVOVR TSVM)]{\includegraphics[width=0.32\linewidth]{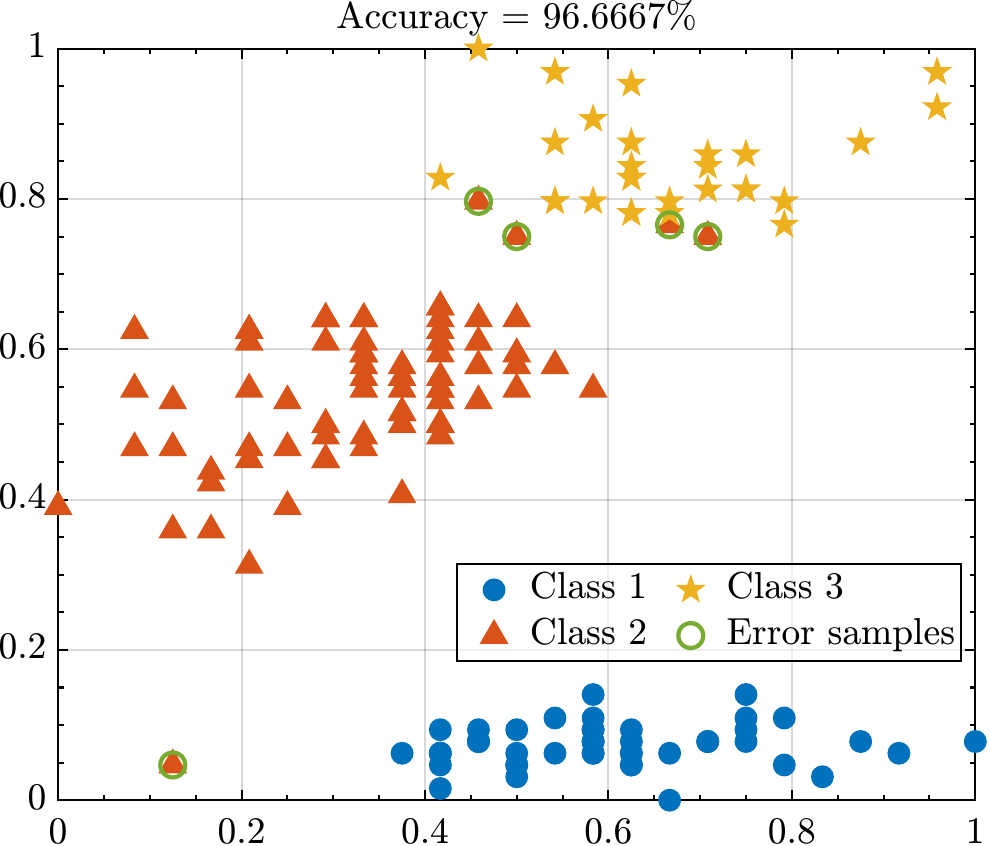} \label{fig_plot_example_pred_tsvm}}
	\subfigure[(1,2)]{\includegraphics[width=0.32\linewidth]{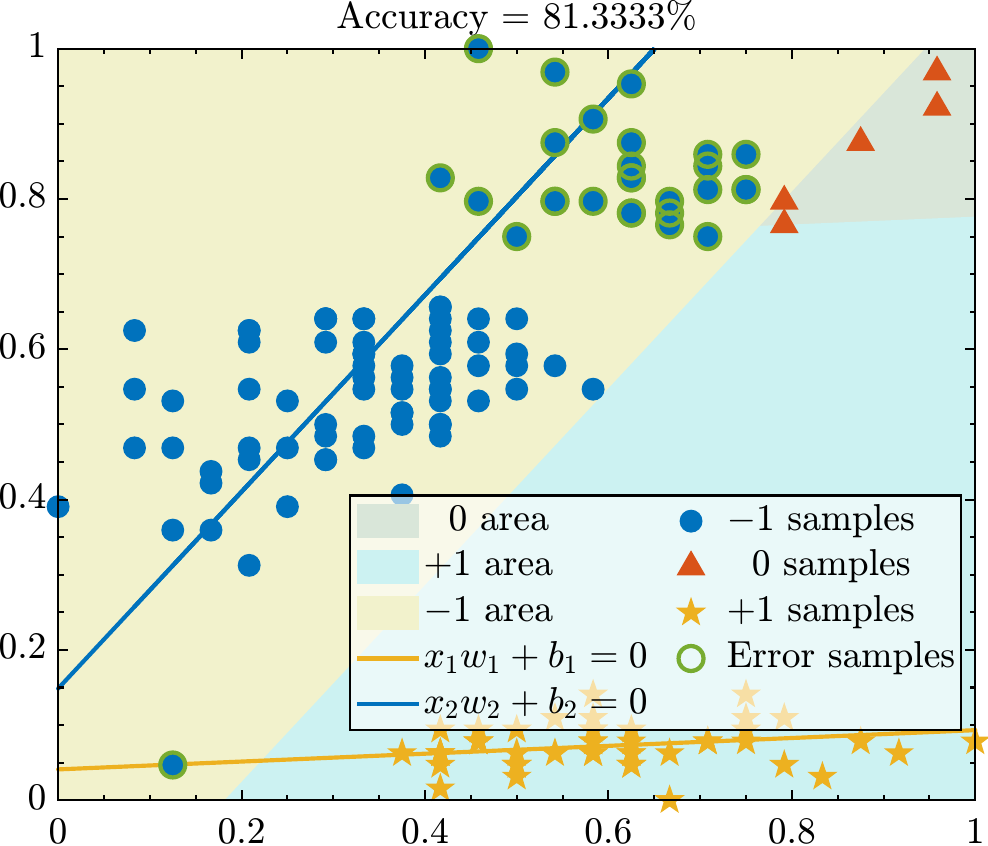} \label{fig_plot_example_res1}}
	\subfigure[(1,3)]{\includegraphics[width=0.32\linewidth]{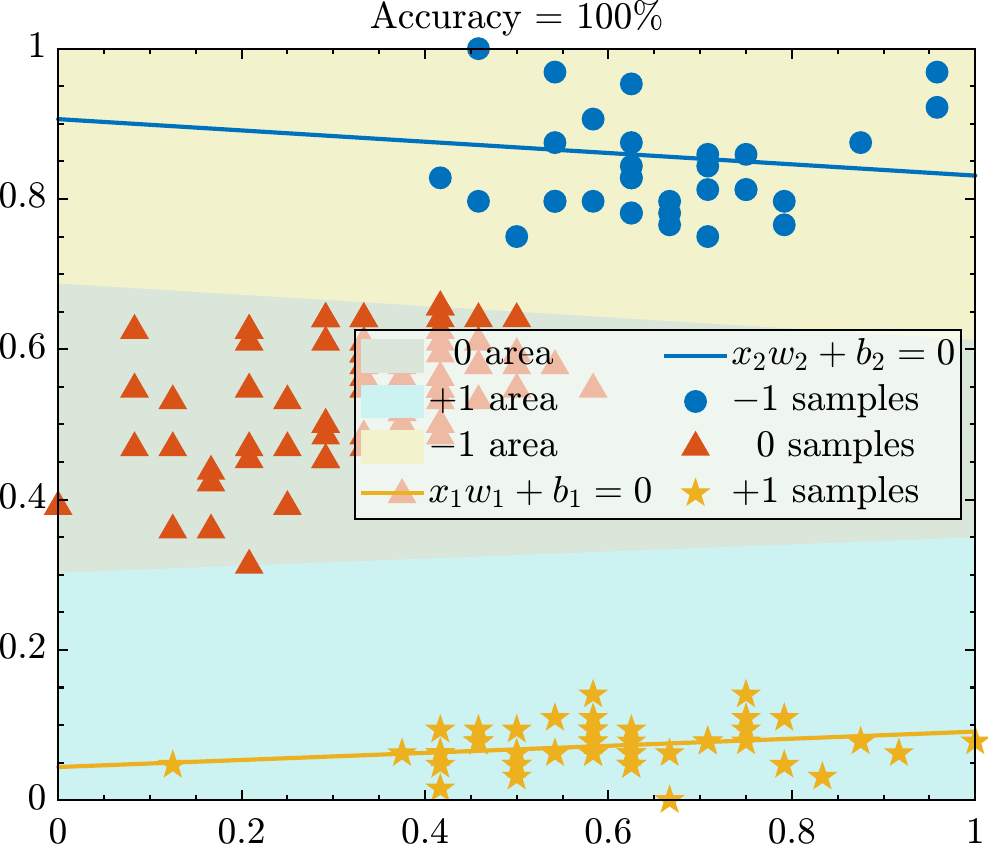} \label{fig_plot_example_res2}}
	\subfigure[(2,3)]{\includegraphics[width=0.32\linewidth]{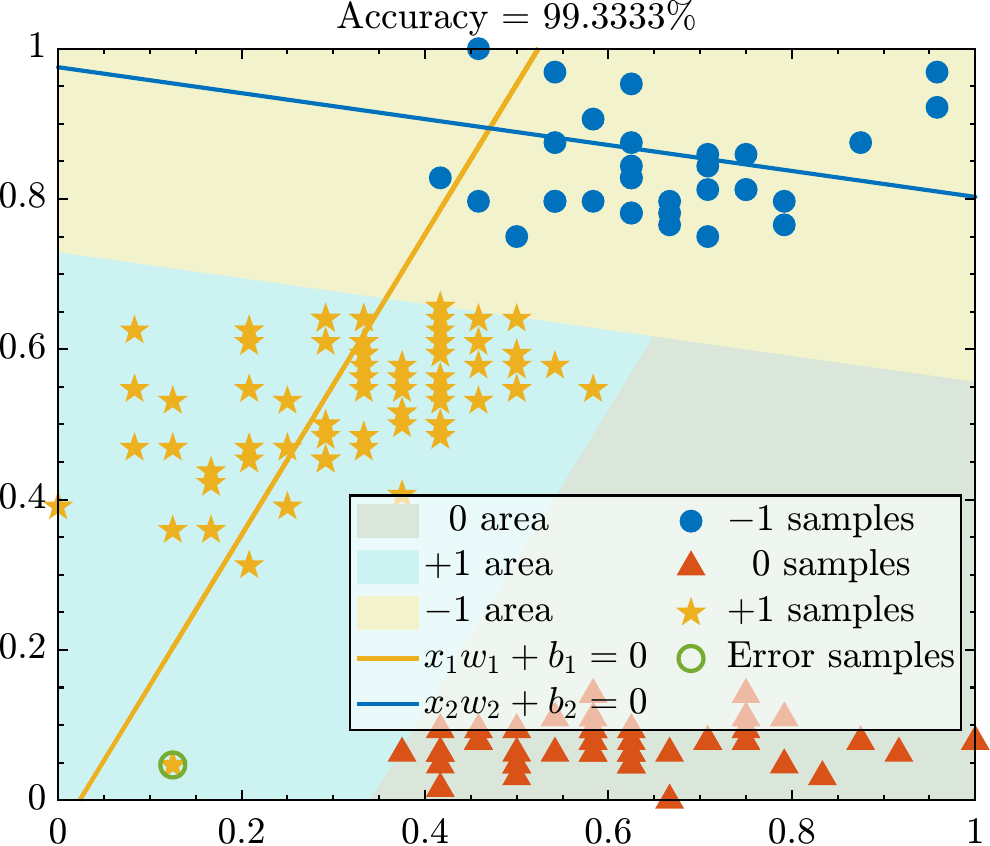} \label{fig_plot_example_res3}}
\caption{Plots of classification results: (a) is the plot of original data with three classes; (b) and (c) are predicted results of ours and OVOVR TSVM; (d) to (f) are predicted results of three combinations (1,2), (1,3) and (2,3), respectively. (Best viewed in color.)}
\label{fig_plot_example}
\end{figure*}

\paragraph{Visualization of classification results}
\hl{To intuitively verify the correctness of our proposed algorithm, we first construct a handcrafted data set, including 3 classes with two-dimensional feature, and the test set is shown in \mbox{\Cref{fig_plot_example_orig}}. Because it contains three classes, we need to evaluate three combinations as mentioned in \mbox{\Cref{sec_algorithm}}, \ie, (1,2), (1, 3) and (2, 3). The corresponding results are depicted in \mbox{\Cref{fig_plot_example_res1,fig_plot_example_res2,fig_plot_example_res3}}, where samples in three classes are pinked into different colored markers and two non-parallel lines are shown in different colors. For example, \mbox{\Cref{fig_plot_example_res3}} illustrates the results of the combination (2,3). Three classes are shown in \mbox{\textcolor{matlabblue}{blue}} circles, \mbox{\textcolor{matlabred}{red}} triangles and \mbox{\textcolor{matlabyellow}{yellow}} pentagrams, labeled in ``$-1$'', ``0'' and ``$+1$''. Two non-parallel lines are shown in \mbox{\textcolor{matlabyellow}{yellow}} and \mbox{\textcolor{matlabblue}{blue}}, which correspond to $w_1 x_1 + b_1 = 0$  and $w_2 x_2 + b_2 = 0$. Two decision boundaries $w_1 x_1 + b_1 > -1 + \epsilon$ and $w_2 x_2 + b_2 < 1 - \epsilon$ divides the whole space to three areas \textit{w.r.t.} three classes. The misclassified samples are marked by a \mbox{\textcolor{matlabgreen}{green}} circle. As shown in \mbox{\Cref{fig_plot_example_res3}}, the $+1$ area contains a misclassified sample and it achieves an accuracy of 99.3333\%. The final results are shown in \mbox{\Cref{fig_plot_example_pred}}, where our proposed algorithm achieves an accuracy of 99.3333\%. However, the original OVOVR TSVM in \mbox{\Cref{fig_plot_example_pred_tsvm}} only achieves an accuracy of 96.6667\%, which is 2.6666\% less than ours.}
\begin{figure*}
\centering
	\subfigure[CV (1,2)]{\includegraphics[width=0.32\linewidth]{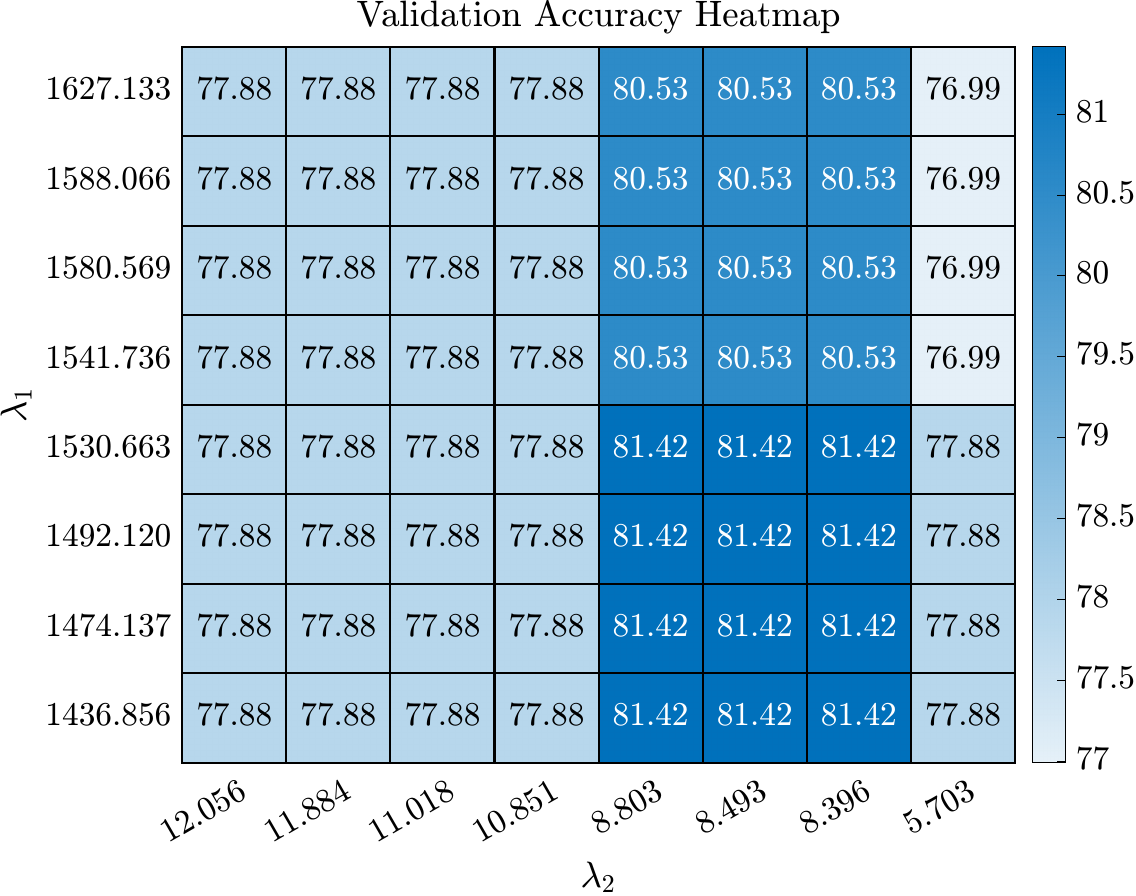}}
	\subfigure[CV (1,3)]{\includegraphics[width=0.32\linewidth]{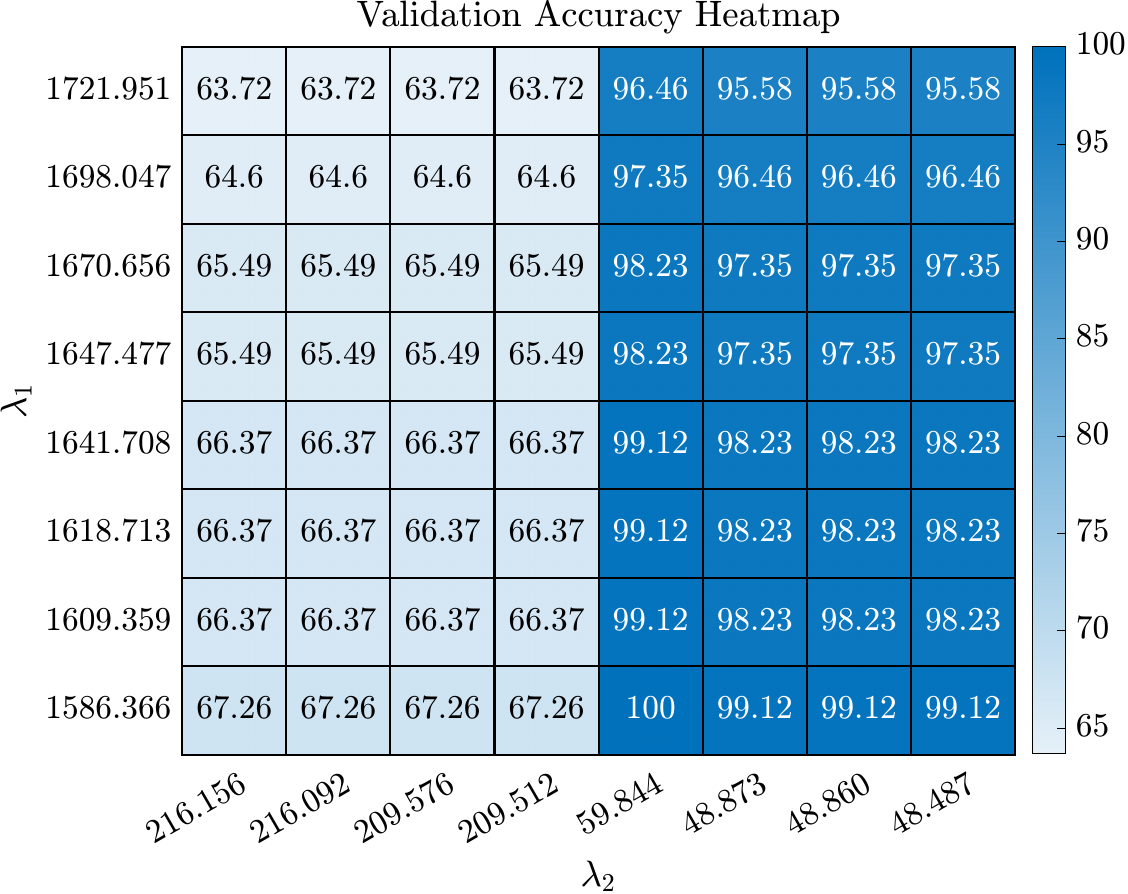}}
	\subfigure[CV (2,3)]{\includegraphics[width=0.32\linewidth]{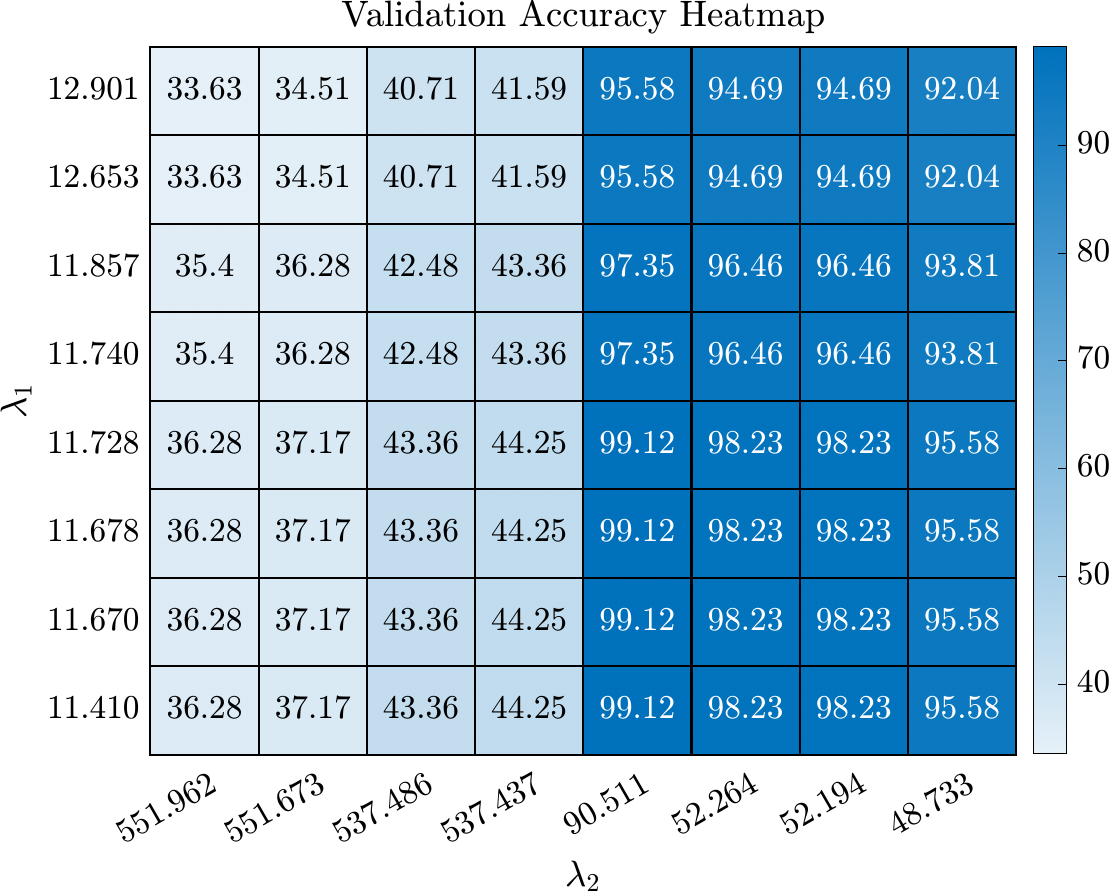}\label{fig_heatmap_example_c}}
\caption{Partial heatmaps of cross-validation accuracy: (a) to (c) are cross-validation heatmaps of three combinations (1,2), (1,3) and (2,3), respectively.}
\label{fig_heatmap_example}
\end{figure*}

\hl{Otherwise, \mbox{\Cref{fig_heatmap_example}} depicts the partial heatmaps of cross-validation accuracy for such three combinations. For example, \mbox{\Cref{fig_heatmap_example_c}} shows the validation heatmap of the combination (2,3), where our proposed algorithm achieves the highest accuracy of 99.12\% on the validation set. Notably, it achieves an accuracy of 99.3333\% on the testing set, indicating the effectiveness of our proposed algorithm.}

\begin{figure*}
	\centering
	\subfigure[$\lambda_1$]{\includegraphics[width=0.30\linewidth, height=0.15\linewidth]{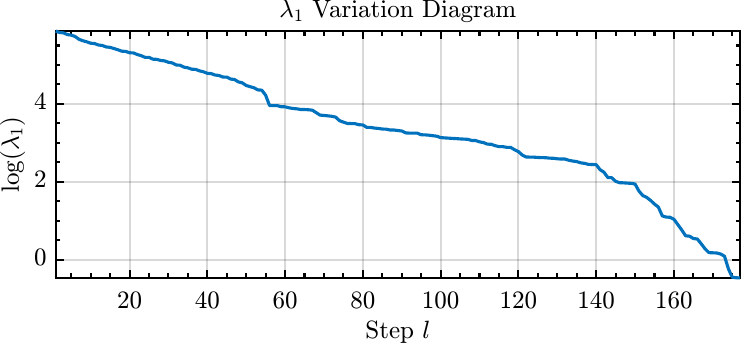} \label{Fig_piecewise_lambda_1}}
	\subfigure[$\bm{\alpha}$]{\includegraphics[width=0.30\linewidth, height=0.15\linewidth]{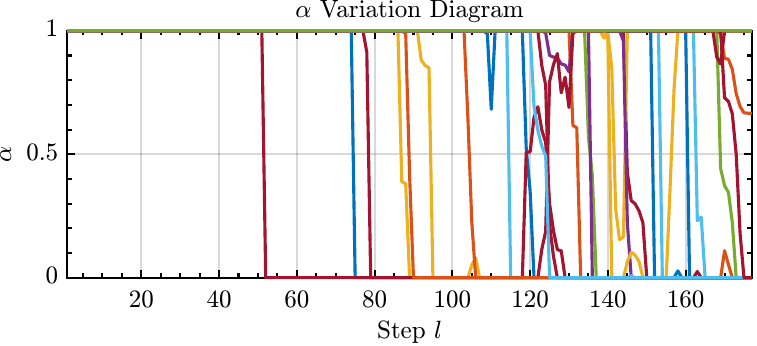} \label{Fig_piecewise_alpha}}
	\subfigure[$\bm{\beta}$]{\includegraphics[width=0.30\linewidth, height=0.15\linewidth]{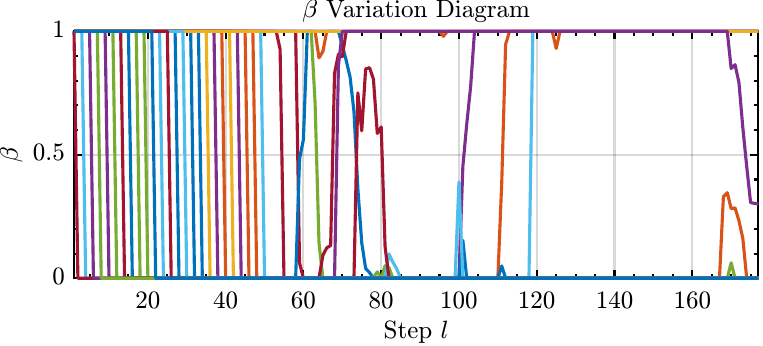} \label{Fig_piecewise_beta}}
	\subfigure[$\bm{f}_1$] { \includegraphics[width=0.30\linewidth, height=0.15\linewidth]{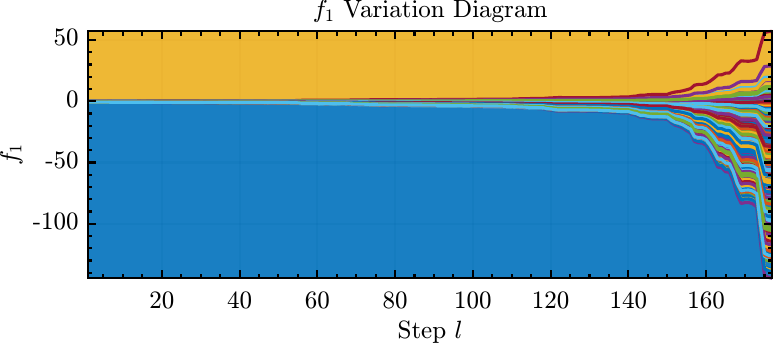} \label{Fig_piecewise_f_1}}
	\subfigure[$\bm{w}_1$] { \includegraphics[width=0.30\linewidth, height=0.15\linewidth]{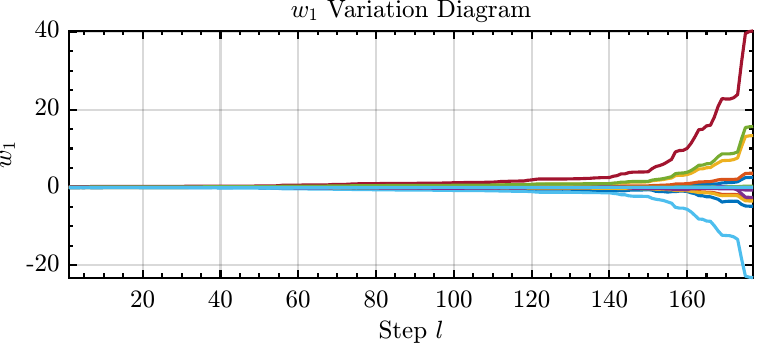} \label{Fig_piecewise_w_1}}
	\subfigure[$b_1$] { \includegraphics[width=0.30\linewidth, height=0.15\linewidth]{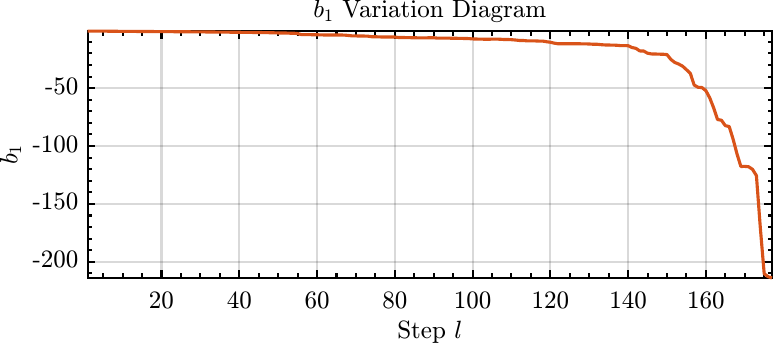} \label{Fig_piecewise_b_1}}
	\caption{An entire solution path of regularization parameters $\lambda_1$ and $\lambda_2$ on data set Wine for the first QPP: (a) to (f) are the plots of $\lambda_1$, $\bm{\alpha}$, $\bm{\beta}$, $\bm{f}_1$, $\bm{w}_1$ and $b_1$ respectively. (Note that $\lambda_1$ is on the $\log$ scale.)}
	\label{Fig_piecewise}
\end{figure*}

\begin{figure*}
	\centering
	\subfigure[$\lambda_2$]{\includegraphics[width=0.30\linewidth, height=0.15\linewidth]{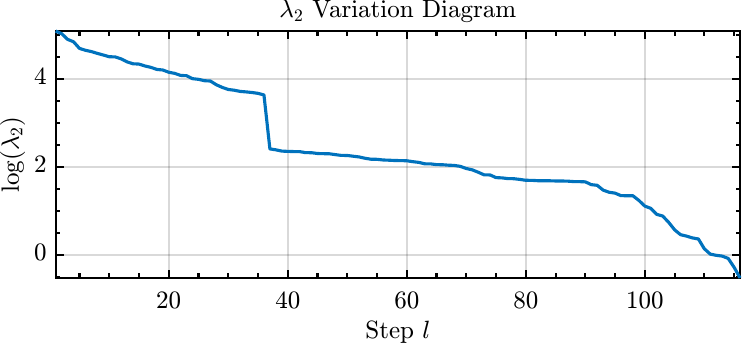} \label{Fig_piecewise_lambda_2}}
	\subfigure[$\bm{\alpha}$]{\includegraphics[width=0.30\linewidth, height=0.15\linewidth]{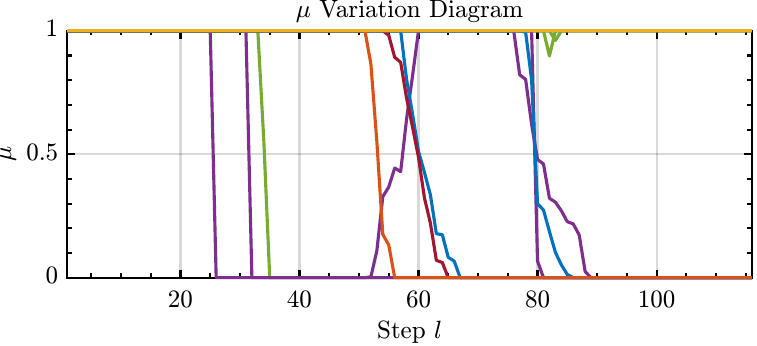} \label{Fig_piecewise_mu}}
	\subfigure[$\bm{\beta}$]{\includegraphics[width=0.30\linewidth, height=0.15\linewidth]{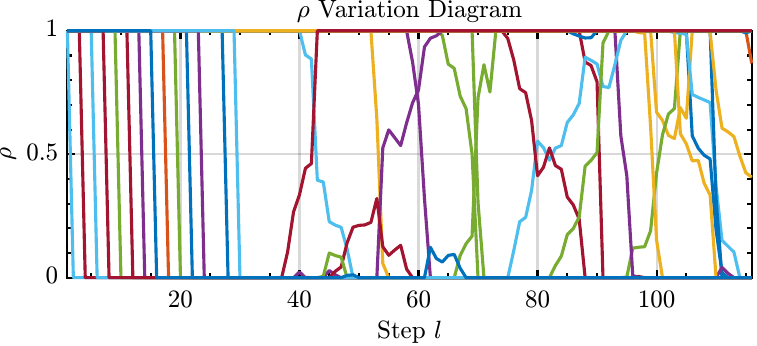} \label{Fig_piecewise_rho}}
	\subfigure[$\bm{f}_2$] { \includegraphics[width=0.30\linewidth, height=0.15\linewidth]{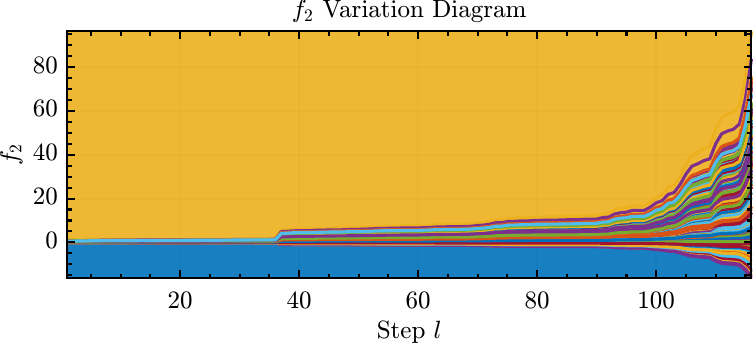} \label{Fig_piecewise_f_2}}
	\subfigure[$\bm{w}_2$] { \includegraphics[width=0.30\linewidth, height=0.15\linewidth]{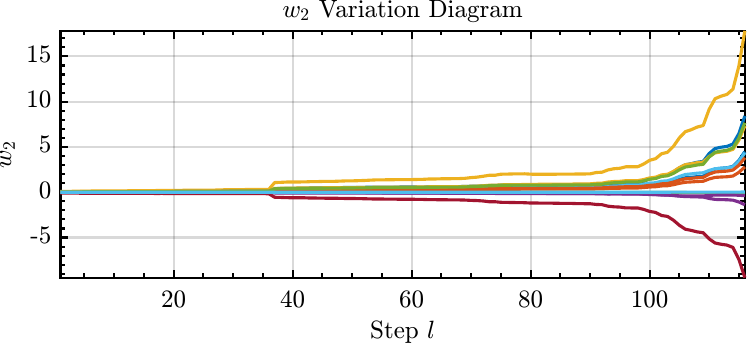} \label{Fig_piecewise_w_2}}
	\subfigure[$b_2$] { \includegraphics[width=0.30\linewidth, height=0.15\linewidth]{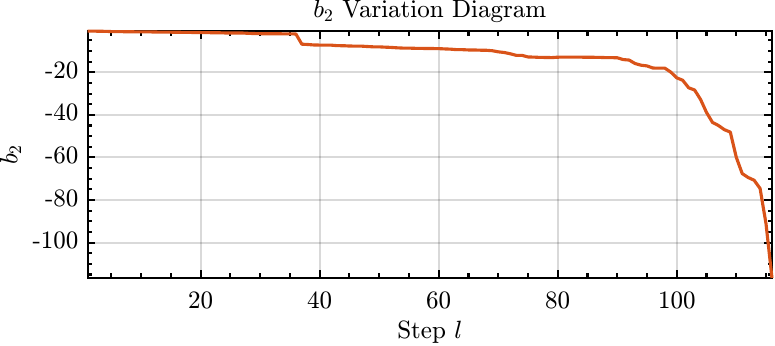} \label{Fig_piecewise_b_2}}
	\caption{An entire solution path on the data set Wine for the second QPP: (a) to (f) are the plots of $\lambda_2$, $\bm{\mu}$, $\bm{\rho}$, $\bm{f}_2$, $\bm{w}_2$ and $b_2$ respectively. (Note that $\lambda_2$ is on the $\log$ scale.)}
	\label{Fig_piecewise_2}
\end{figure*}

\paragraph{Visualization of piecewise linear solution path}
\Cref{Fig_piecewise,Fig_piecewise_2} show the entire solution path of the regularization parameter $\lambda_1$ for the QPP \eqref{eqn3} and $\lambda_2$ for the QPP \eqref{eqn2_t} on the data set Wine, where the regularization parameters are on the $log$ scale.
\Cref{Fig_piecewise_lambda_1,Fig_piecewise_lambda_2} depict the variation digram of regularization parameters $\lambda_1$ and $\lambda_2$, where it can be seen that the regularization parameters are reduced continuously to 0 with the step.
\Cref{Fig_piecewise_alpha,Fig_piecewise_beta,Fig_piecewise_mu,Fig_piecewise_rho} are entire solutions of the Lagrangian multipliers $\bm{\alpha}$, $\bm{\beta}$, $\bm{\mu}$ and $\bm{\rho}$ respectively. For example, combining \Cref{Fig_piecewise_alpha} with \Cref{Fig_piecewise_lambda_1}, it can be seen that the Lagrangian multipliers $\bm{\alpha}$ are piecewise linear \textit{w.r.t.} the regularization parameter $\lambda_1$;
combining \Cref{Fig_piecewise_mu} with \Cref{Fig_piecewise_lambda_2}, it can be seen that the Lagrangian multipliers $\bm{\mu}$ are piecewise linear \textit{w.r.t.} the regularization parameter $\lambda_1$.
Therefore, the piecewise linear theory established in \Cref{theorem1,theorem2} can be experimentally verified.

Additionally, \Cref{Fig_piecewise,Fig_piecewise_2} also show the entire solution path of $\bm{f}$, $\bm{w}$ and $b$ \textit{w.r.t.} two sub-optimization problems, respectively. For example, it can be seen from \Cref{Fig_piecewise_f_1,Fig_piecewise_f_2} that the function values are distributed in different intervals. As mentioned in \Cref{sec_algorithm}, when $f_1 > -1 + \epsilon$ in \Cref{Fig_piecewise_f_1}, we can obtain the $+1$ samples. In the similar way, when $f_2 < 1 - \epsilon$ in \Cref{Fig_piecewise_f_2}, we can obtain the $-1$ samples. Combining \Cref{Fig_piecewise_f_1} with \Cref{Fig_piecewise_f_2}, we can obtain the rest samples.

\subsubsection{Prediction Accuracy Results}

\begin{figure*}
	\centering
	
	\subfigure[]{\includegraphics[width=0.3\linewidth, height=0.24\linewidth]{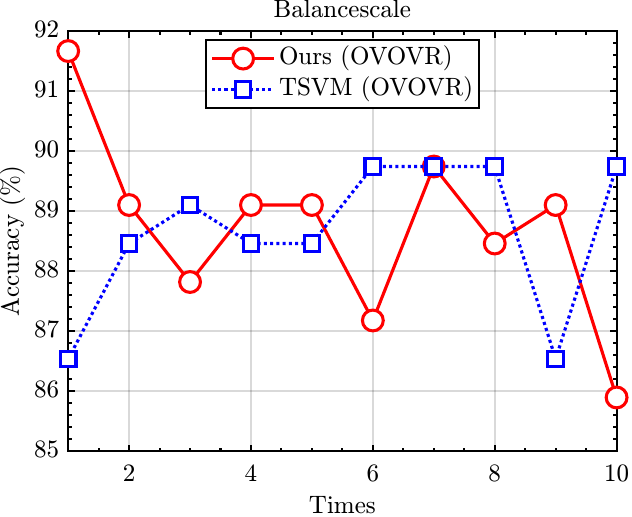} \label{Fig_acc_2}}
	\subfigure[]{\includegraphics[width=0.3\linewidth, height=0.24\linewidth]{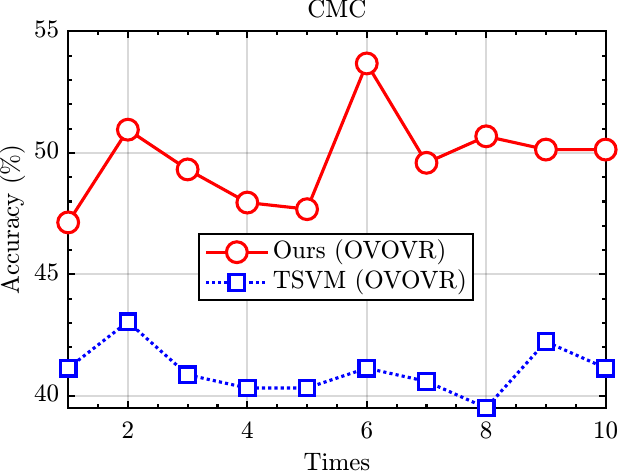} }
	\subfigure[]{\includegraphics[width=0.3\linewidth, height=0.24\linewidth]{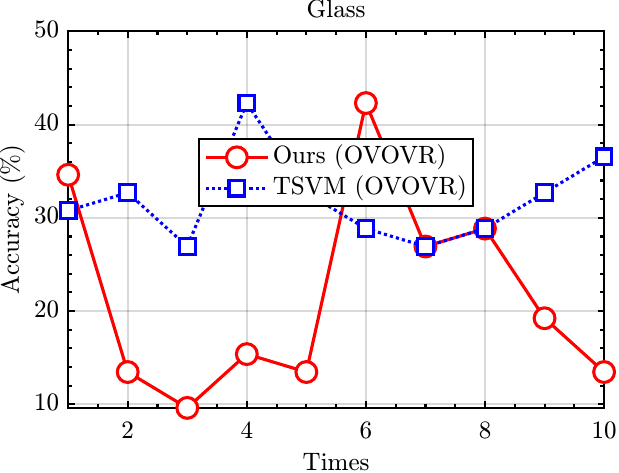} }
	\subfigure[]{\includegraphics[width=0.3\linewidth, height=0.24\linewidth]{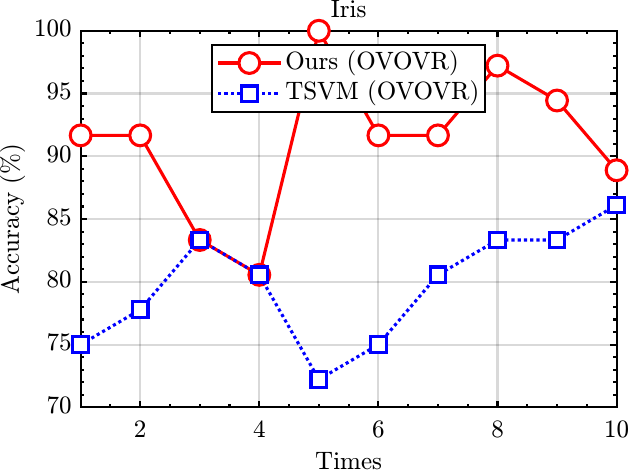} }
	\subfigure[]{\includegraphics[width=0.3\linewidth, height=0.24\linewidth]{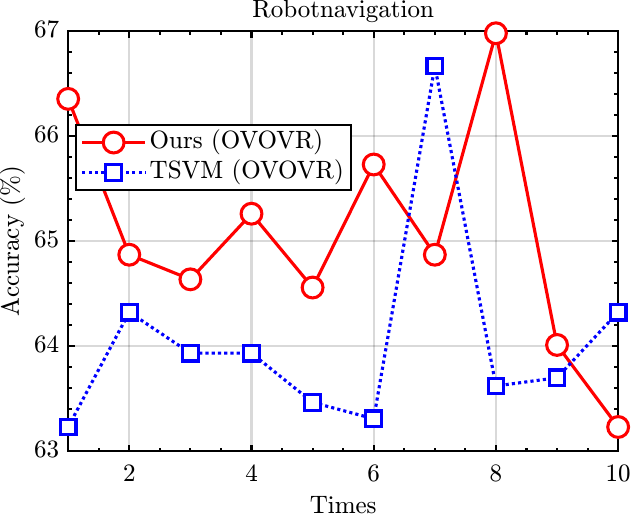} }
	\subfigure[]{\includegraphics[width=0.3\linewidth, height=0.24\linewidth]{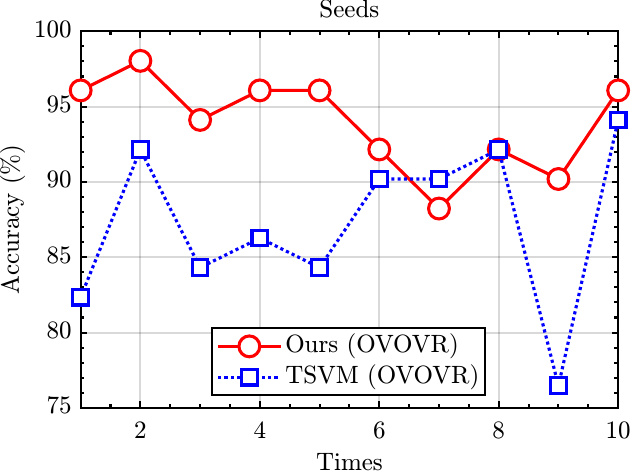} }
	\subfigure[]{\includegraphics[width=0.3\linewidth, height=0.24\linewidth]{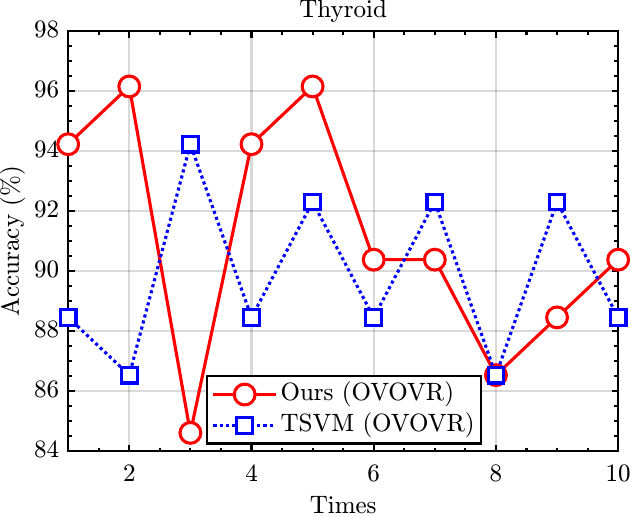} }
	\subfigure[]{\includegraphics[width=0.3\linewidth, height=0.24\linewidth]{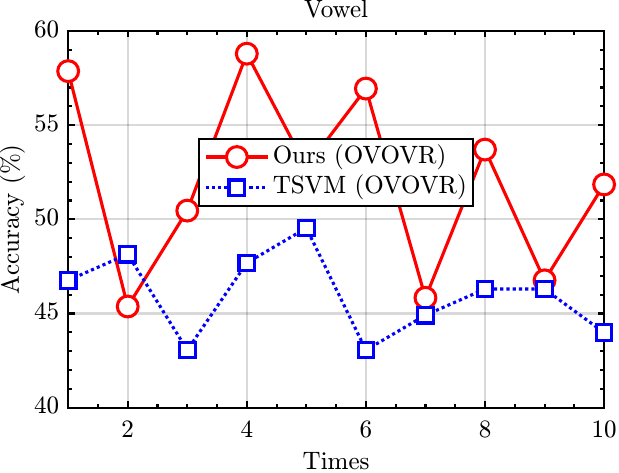} }
	\subfigure[]{\includegraphics[width=0.3\linewidth, height=0.24\linewidth]{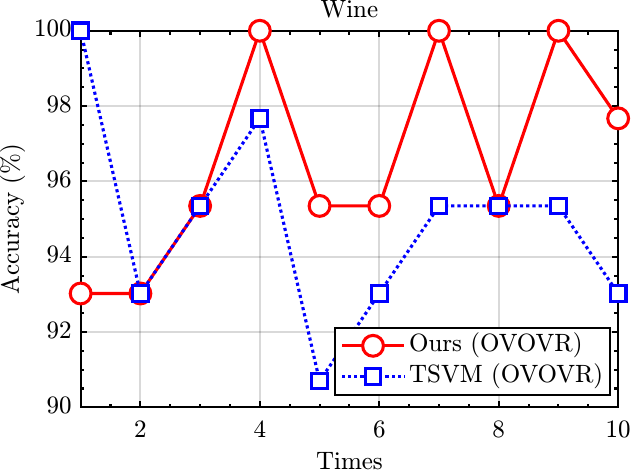} \label{Fig_acc_12}}

	\caption{Prediction accuracy plots of ten times repeated experiments using ours and grid search method: (a) to (i) are prediction accuracy plots on each data set respectively; the solid red line with circles and the dotted line with squares denote the results of ours and the grid search method.}
	\label{Fig_acc}
\end{figure*}

\paragraph{Ours vs grid search method} Faced with the parameter tuning problem, one of the main strengths of the proposed algorithm is that we can search for an entire  solution path in the parameter space whereas the grid search method can only find out limited solutions. Therefore, our prediction accuracy results on the same data set are better than the grid search method in most scenarios. Adopting the same data set division strategy  on nine different data sets, we have tested the proposed algorithm and grid search method and the corresponding results are shown in \Cref{Fig_acc}. \Cref{Fig_acc_2} to \Cref{Fig_acc_12} are plots on nine data sets respectively. In each plot, the solid red line with circle markers and the blue dotted line with square markers denote ours and the grid search method respectively.  Note that we have repeated ten times on each data set using different data set divisions in order to ensure the reliability of the experiment. As can be seen in \Cref{Fig_acc}, our prediction results are much better than the grid search method on several data sets such as CMC, Iris, Seeds, Robotnavigation and Vowel. For the other data sets, the prediction results for the two methods are about the same.


\paragraph{Other baselines} 
For the multi-classification problem, TSVM can be combined with different strategies. Note that the proposed algorithm is based on the OVOVR strategy. In this work, we compared our algorithm with OVO TSVM and OVR TSVM. 
Furthermore, SVM can be combined with OVO and OVR strategies to solve the multi-classification problems. Therefore,
we have also tested OVO SVM and OVR SVM. The detailed results are elaborated in \Cref{Tab_acc}.
Compared with other algorithms, the prediction accuracy of the proposed algorithm is better than other methods on dataset CMC, Iris, Robotnavigation, Seeds and Wine. For the data set Iris, Seeds and Wine, the number of samples in different categories is roughly the same. Therefore, the prediction accuracy is very high. However, the prediction accuracies of different algorithms are quite low for the data set CMC and Glass. The reason is that this data set may be linearly indivisible, while we use the \hl{linear kernel} for all algorithms in this work.
For data set Glass and Vowel with more than 3 classes, the prediction accuracy of ours and TSVM (OVOVR) are lower than other algorithms. It is demonstrated that the OVOVR strategy is slightly inferior compared with other strategies. 

\hl{Although the prediction accuracy of our algorithm is not as good as that of other algorithms on some data sets, the prediction accuracy of our algorithm is generally superior to that of TSVM with the same OVOVR strategy as can be seen from \mbox{\Cref{Fig_acc}}, indicating the effectiveness of our proposed solution path algorithm. We highlight that the main advantage of our algorithm is its low computational overhead and complexity with comparable results, while the performance lifting will be our future work. However, we have to admit that there are some limitations to the OVOVR strategy, \eg, cross-validation among different combinations is very time-consuming and the local optimal solution combination cannot always achieve the global optimal.}

\begin{table*}
\tiny
	\caption{Average prediction accuracy (\%) of different algorithms on each data sets.}
	\label{Tab_acc}
	\centering
		\begin{tabular}{ccccccc}
		\toprule
		\multirow{2}{*}{Data set} & 
		\multicolumn{1}{c}{\tabincell{c}{Ours\\(OVOVR)} } &
		\multicolumn{1}{c}{\tabincell{c}{TSVM\\(OVOVR)} } &
		\multicolumn{1}{c}{\tabincell{c}{TSVM\\(OVR)} } &
		\multicolumn{1}{c}{\tabincell{c}{TSVM\\(OVO)}} &
		\multicolumn{1}{c}{\tabincell{c}{SVM\\(OVR)}} & 
		\multicolumn{1}{c}{\tabincell{c}{SVM\\(OVO)}} \\ \cline{2-7}
		&
		Acc, $\pm$ Std. &
		Acc, $\pm$ Std. &
		Acc, $\pm$ Std. &
		Acc, $\pm$ Std. &
		Acc, $\pm$ Std. &
		Acc, $\pm$ Std. \\
		\midrule
 Balancescale & 88.72 $\pm$ 1.55 & 88.65 $\pm$ 1.25 & 91.92 $\pm$ 2.44 & 88.21 $\pm$ 1.14 & 88.40 $\pm$ 1.61 & 91.47 $\pm$ 1.84 \\
 CMC & 49.73 $\pm$ 1.90 & 41.04 $\pm$ 1.00 & 45.69 $\pm$ 1.07 & 43.92 $\pm$ 1.53 & 48.45 $\pm$ 1.96 & 51.36 $\pm$ 1.95 \\
 Glass & 21.73 $\pm$ 10.88 & 31.92 $\pm$ 4.73 & 17.69 $\pm$ 7.73 & 32.69 $\pm$ 2.56 & 24.62 $\pm$ 4.42 & 39.23 $\pm$ 4.81 \\
 Iris & 88.61 $\pm$ 7.23 & 79.72 $\pm$ 4.55 & 75.00 $\pm$ 7.52 & 96.39 $\pm$ 2.94 & 93.89 $\pm$ 4.10 & 98.33 $\pm$ 1.43 \\
 Robotnavigation & 65.05 $\pm$ 1.09 & 64.05 $\pm$ 0.99 & 60.43 $\pm$ 2.81 & 60.38 $\pm$ 3.03 & 73.32 $\pm$ 1.34 & 75.07 $\pm$ 1.05 \\
 Seeds & 93.92 $\pm$ 3.13 & 87.25 $\pm$ 5.49 & 92.16 $\pm$ 2.77 & 92.75 $\pm$ 1.61 & 92.35 $\pm$ 3.26 & 91.76 $\pm$ 4.01 \\
 Thyroid & 91.15 $\pm$ 3.97 & 89.81 $\pm$ 2.73 & 83.27 $\pm$ 2.41 & 85.00 $\pm$ 4.33 & 95.96 $\pm$ 2.79 & 97.31 $\pm$ 3.03 \\
 Vowel & 52.04 $\pm$ 4.95 & 45.97 $\pm$ 2.19 & 54.07 $\pm$ 1.82 & 53.56 $\pm$ 1.84 & 42.08 $\pm$ 13.42 & 81.67 $\pm$ 2.24 \\
 Wine & 96.51 $\pm$ 2.74 & 94.88 $\pm$ 2.64 & 96.28 $\pm$ 2.94 & 94.65 $\pm$ 3.30 & 95.81 $\pm$ 2.14 & 95.58 $\pm$ 3.19 \\
 \bottomrule
		\end{tabular}
\end{table*}

\subsubsection{Training Time Comparison} 

The average training time of different algorithms is summarized in \Cref{Tab_time}.
In \Cref{Tab_time}, we also list the average number of starting events of a solution path. Furthermore, \Cref{Tab_event} elaborates on the average number of starting events on each data set.
The number of starting events reflects the size of the corresponding solution path. To save time, we just evaluate the training time of one solution for different algorithms. 
The proposed algorithm can be used to solve the entire solution path about regularization parameters. From \Cref{Tab_time}, the training time of ours is quite less than that of the others. 
Since we do not need to solve any QPP to obtain the entire solution path, it can be seen from \Cref{Tab_time} that the time to solve QPP on the same data set is much longer than that of ours, the linear solving time. Therefore, the main factor restricting the training time is solving QPPs. The time to solve a QPP each time also depends on the size of the data set, and for high-dimensional data sets, the time to solve a QPP may even exceed a few hours. Therefore, the time of solving QPPs is used to measure the computational cost of different algorithms in this work.

\begin{table*}
\tiny
	\caption{Average training time (s) of different algorithms on each data sets.}
	\label{Tab_time}
	\centering
	\begin{tabular}{ccccccccccc}
	\toprule
	\multirow{2}{*}{Data set} &
	\multicolumn{3}{c}{\# Average starting events} &
	\multirow{2}{*}{Time $^ {\rm{a}}$ (s)} &
	\multicolumn{6}{c}{Training time of each solution (s)}  \\\cline{2-4} \cline{6-11}
	&
	QPP \eqref{eqn3} & QPP \eqref{eqn2_t} & Total &
	 & \tabincell{c}{Ours $^ {\rm{b}}$\\(OVOVR)} & \tabincell{c}{TSVM\\(OVOVR)} & \tabincell{c}{TSVM\\(OVR)} & \tabincell{c}{TSVM\\(OVO)}  & \tabincell{c}{SVM\\(OVR)} & \tabincell{c}{SVM\\(OVO)} \\
	\midrule
 Balancescale & 517 & 731 & 1247 & 1.3149 & 0.0011 & 3.4525 & 0.8231 & 1.2516 & 0.0341 & 0.0331 \\
 CMC & 1600 & 1053 & 2653 & 19.6988 & 0.0074 & 8.3385 & 4.5832 & 5.6763 & 2.4142 & 1.2867 \\
 Glass & 2295 & 1587 & 3882 & 2.9809 & 0.0008 & 2.1671 & 0.4313 & 1.0795 & 0.0638 & 0.1211 \\
 Iris & 122 & 123 & 245 & 0.7215 & 0.0029 & 0.1368 & 0.1049 & 0.1196 & 0.0266 & 0.0273 \\
 Robotnavigation & 925 & 1000 & 1925 & 80.2642 & 0.0417 & 466.2647 & 239.3250 & 215.2090 & 21.2384 & 5.9840 \\
 Seeds & 2880 & 1961 & 4841 & 0.5551 & 0.0001 & 0.1492 & 0.1428 & 0.1674 & 0.0278 & 0.0279 \\
 Thyroid & 2974 & 2131 & 5105 & 0.4211 & 0.0001 & 0.2541 & 0.1876 & 0.2320 & 0.0762 & 0.0299 \\
 Vowel & 3920 & 2957 & 6877 & 17.5491 & 0.0026 & 28.0308 & 7.9994 & 13.3788 & 53.1428 & 17.6766 \\
 Wine & 178 & 209 & 387 & 0.8376 & 0.0022 & 0.1640 & 0.2616 & 0.1128 & 2.1981 & 0.9186 \\
 	\bottomrule
	\end{tabular}
	\begin{threeparttable}
	\begin{tablenotes}
		\item[a] Average training time of entire solution for ours.
		\item[b] Average training time of each solution for ours.
	\end{tablenotes}
	\end{threeparttable}
\end{table*}	

\begin{table*}
\tiny
	\caption{Average number of starting events on each data sets.}
	\label{Tab_event}
	\centering
	\begin{tabular}{cccccccccc}
	\toprule
	Data set & 
	Event 1 & Event 2 & Event 3 & Event 4 &
	Event 5 & Event 6 & Event 7 & Event 8 &  Sum\\
	\midrule
 Balancescale &  53 & 203 & 206 & 61 & 85 & 269 & 275 & 95 & 1247 \\
 CMC &  68 & 389 & 399 & 90 & 221 & 587 & 745 & 154 & 2653 \\
 Glass &  88 & 491 & 500 & 116 & 331 & 957 & 1133 & 267 & 3882 \\
 Iris &  7 & 53 & 51 & 9 & 7 & 55 & 53 & 10 & 245 \\
 Robotnavigation &  71 & 321 & 337 & 72 & 129 & 425 & 464 & 106 & 1925 \\
 Seeds &  136 & 680 & 685 & 169 & 374 & 1147 & 1335 & 315 & 4841 \\
 Thyroid &  147 & 746 & 751 & 179 & 384 & 1192 & 1380 & 327 & 5105 \\
 Vowel &  185 & 928 & 936 & 227 & 499 & 1722 & 1927 & 453 & 6877 \\
 Wine &  25 & 65 & 67 & 28 & 25 & 74 & 71 & 32 & 387 \\
 	\bottomrule
	\end{tabular}
\end{table*}	

\subsection{Discussion}
In this section, we discuss the performance of the proposed algorithm in terms of the computational overhead and the time complexity, respectively.

\paragraph{Computational Overhead} Because the maximum regularization parameter $\lambda$ is reduced from 1000, and the step size of the regularization parameter $\lambda$ is $\Delta \lambda = 0.1$ when the grid search method is adopted. 
Therefore, we need to solve $10 ^ {8}$ QPPs in each combination $(K_{i}, K_{j})$ for parameter tuning. In addition, we need to solve $\mathcolorbox{myhlcolortwo}{K(K - 1) \times 10 ^ {8}}$ QPPs to figure out the solution path. Due to the limitation of computing power, the grid search method cannot find the entire regularized solution path. 
Fortunately, the proposed solution path algorithm can expand the solution space of the regularized solution path to $(0, +\infty)$. 
Furthermore, no QPP is required for the proposed solution path algorithm. 
Since the training time of ours is linear, which is much less than that of solving QPP.
Obviously, compared with the grid search method, the computational cost of the proposed algorithm will be greatly reduced.
It can be seen that the proposed solution path algorithm has a significant advantage in parameter tuning.

\paragraph{Time Complexity}
\hl{Since \mbox{\Cref{algorithm1}} needs to solve linear equations of size $n_B + n_C$, its time complexity is $\mathcal{O}(\bar{n}^2)$ at least where $\bar{n}$ denotes the average sample size.
According to \mbox{\citep{Hastie2004}}, the time complexity of \mbox{\Cref{algorithm2}} is $\mathcal{O}(c\bar{n}^2\bar{m} + \bar{n}\bar{m}^2)$, where $\bar{m}$ is the average size of $\mathcal{E}_{\mathrm{B}}$ and $\mathcal{E}_{\mathrm{C}}$ and $c$ is a small number. 
In summary, the time complexity of the whole algorithm is proportional to the square of the data size.
Additionally, the total computation burden of the entire solution path algorithm is similar to that of a single OVOVR TSVM fit. 
For example, \mbox{\cite{chen2017multiple}} does not pay much attention to the complexity of parameter tuning, thus the classification accuracy can be guaranteed. They need to solve $K$ QPPs to obtain $K$ hyperplanes for the multi-classification problem, which is more expensive than ours.
For the grid search method, we need to fit the OVOVR TSVM $n_{\mathrm{grid}}$ times, and the corresponding time complexity is also $n_{\mathrm{grid}}$ times of OVOVR TSVM fits, where $n_{\mathrm{grid}}$ is the granularity of the grid, \textit{e.g.}, $n_{\mathrm{grid}}$ is equal to $2 \times 10^4$ as analyzed above in this work. 
Therefore, the solution path algorithm can greatly reduce the computational burden of parameter adjustment,
with up to four orders of magnitude speed-up for the computational complexity compared with the grid search method. Furthermore, the convergence of solution path algorithm can be guaranteed by \mbox{\cite{Hastie2004}}.}

\section{Conclusion} \label{Conclusion}

In this work, the twin multi-class SVM with the OVOVR strategy is studied and its fast regularization parameter tuning algorithm is developed. The solutions of the two sub-optimization problems are proved to be piecewise linear on the regularization parameters, and the entire regularized solution path algorithm is developed accordingly. The simulation results on UCI data sets show that the Lagrangian multipliers are piecewise linear \textit{w.r.t.} the regularization parameters of the two sub-models, which lays a foundation for further selecting regularization parameters and makes the generalization performance of the twin multi-class support vector machine better. 
It should be noted that no QPP is involved in the proposed algorithm, thus  sharply reducing the computational cost.

\section*{Acknowledgments}
This work was supported by the Natural Science Foundation of China (61203293, 61702164, 31700858), Scientific and Technological Project of Henan Province (162102310461, 172102310535), Foundation of Henan Educational Committee (18A520015).

\bibliographystyle{model5-names}
\bibliography{TwinMultiPath_R3}

\appendix

\end{document}